\documentclass[10pt]{article} % For LaTeX2e
\usepackage[preprint]{tmlr}

\usepackage{amsmath,amsfonts,bm}

\newcommand{\blue}[1]{{\color{blue}#1}}

\newcommand{\cmark}{\ding{51}}%
\definecolor{darkgreen}{RGB}{0,100,0}  
\newcommand{\xmark}{\textcolor{red}{\ding{55}}}  
\newcommand{\midruletight}{\aboverulesep = 0mm \belowrulesep = 0mm}
\newcommand{\eg}{\textit{e}.\textit{g}.}
\newcommand{\ie}{\textit{i}.\textit{e}.}

\def\eqref#1{equation~\ref{#1}}
\def\1{\bm{1}}

\DeclareMathAlphabet{\mathsfit}{\encodingdefault}{\sfdefault}{m}{sl}
\SetMathAlphabet{\mathsfit}{bold}{\encodingdefault}{\sfdefault}{bx}{n}

\usepackage{hyperref}       % hyperlinks
\usepackage{url}            % simple URL typesetting
\usepackage{booktabs}       % professional-quality tables
\usepackage{amsfonts}       % blackboard math symbols
\usepackage{nicefrac}       % compact symbols for 1/2, etc.
\usepackage{microtype}      % microtypography
\usepackage{xcolor}         % colors
\usepackage{amsmath}
\usepackage{multirow}
\usepackage{caption}
\usepackage{makecell}
\usepackage{rotating}
\usepackage{graphicx}
\usepackage{svg}
\usepackage{float}
\usepackage{subcaption}
\usepackage{pifont}
\usepackage{xcolor}
\usepackage{float}
\usepackage[inline]{enumitem}
\usepackage{placeins}
\usepackage{wrapfig}
\usepackage{xspace}
\usepackage{caption}
\usepackage{placeins}

\title{\textsc{VidHal}: Benchmarking Temporal Hallucinations \\in Vision LLMs}

\author{\name Wey Yeh Choong \email \href{mailto:c.weyyeh@u.nus.edu}{c.weyyeh@u.nus.edu} \\
      \addr National University of Singapore
      \AND
      \name Yangyang Guo\thanks{Corresponding author.} \email \href{mailto:guoyang.eric@gmail.com}{guoyang.eric@gmail.com} \\
      \addr National University of Singapore
      \AND
      \name Mohan Kankanhalli \email \href{mailto:mohan@comp.nus.edu.sg}{mohan@comp.nus.edu.sg}\\
      \addr National University of Singapore
}

\def\month{04}  % Insert correct month for camera-ready version
\def\year{2026} % Insert correct year for camera-ready version
\def\openreview{\url{https://openreview.net/forum?id=7ccWCDbdM1}} % Insert correct link to OpenReview for camera-ready version

\begin{document}

\maketitle

\begin{abstract}
Vision Large Language Models (VLLMs) are widely acknowledged to be prone to hallucinations. Existing research addressing this problem has primarily been confined to image inputs, with sparse exploration of their video-based counterparts. Furthermore, current evaluation methods fail to capture nuanced errors in generated responses, which are often exacerbated by the rich spatiotemporal dynamics of videos. To address these two limitations, we introduce \textsc{VidHal}, a benchmark specially designed to evaluate video-based hallucinations in VLLMs. \textsc{VidHal} is constructed by bootstrapping video instances across a wide range of common temporal aspects. A defining feature of our benchmark lies in the careful creation of captions representing varying levels of hallucination associated with each video. To enable fine-grained evaluation, we propose a novel caption ordering task requiring VLLMs to rank captions by hallucinatory extent. We conduct extensive experiments on \textsc{VidHal} and comprehensively evaluated a broad selection of models, including both open-source and proprietary ones such as GPT4.1 and Gemini 2.5. Our results uncover significant limitations in existing VLLMs regarding video-based hallucination generation. Through our benchmark, we aim to inspire further research on I) holistic understanding of VLLM capabilities, particularly regarding hallucination, and II) advancing VLLMs to alleviate this problem. Our \textsc{VidHal} dataset and evaluation code are publicly available at \url{https://github.com/Lookuz/VidHal}.
\end{abstract} 
\section{Introduction}
\label{sec:intro}
Building on the advancements of Large Language Models (LLMs), Vision LLMs (VLLMs) have recently gained significant attention. Models such as LLaVA~\citep{liu2023visualinstructiontuning, liu2024improvedbaselinesvisualinstruction} have shown impressive performance across various visual understanding tasks involving both images and videos. Despite their potential, VLLMs are notably prone to hallucinations, where generated responses appear plausible but contradict visual context~\citep{bai2024hallucinationmultimodallargelanguage, hallucinationisinevitablexu2024}. This problem significantly compromises the reliability of VLLMs, hindering their practical use in real-world applications.

To tackle this challenge, some methods propose to leverage post-hoc techniques such as contrastive decoding~\citep{leng2023mitigatingobjecthallucinationslarge, zhu2024ibdalleviatinghallucinationslarge, m3id, zhuang2024gameontree} and attention calibration~\citep{huang2024operaalleviatinghallucinationmultimodal, ma2023vistallamareliablevideonarrator, liu2024paying, lessismore, xuan2024damro, zhoumitigating2024, xing2024mitigating}. Other efforts have been devoted to the evaluation of hallucinations in VLLMs. For example, CHAIR~\citep{rohrbach2019objecthallucinationimagecaptioning} initially studies object-based hallucination evaluation with the aid of the image captioning task. Subsequent studies~\citep{li2023evaluatingobjecthallucinationlarge, liu2024phdpromptedvisualhallucination, kaul2024throneobjectbasedhallucinationbenchmark, hallupi} instead harness paired \textlangle\textit{positive, hallucinatory}\textrangle\xspace questions to probe such hallucinations. Additionally, MMHalBench~\citep{sun2024aligningmmhalbench} and AMBER~\citep{wang2024amberllmfreemultidimensionalbenchmark} expand beyond object-based evaluations by constructing benchmarks that cover attribute and relationship hallucinations.

Unlike their image-based counterparts, video hallucinations pose unique challenges primarily due to the intricate spatiotemporal dynamics of videos~\citep{ fu2024videommefirstevercomprehensiveevaluation, liu2024tempcompassvideollmsreally, ning2023videobenchcomprehensivebenchmarktoolkit}. In particular, video-specific temporal aspects, such as movement direction and chronological order of events, are especially concerning for video-based VLLMs. Furthermore, the richness of video content necessitates a finer-grained understanding, making VLLMs more vulnerable to nuanced hallucinations. Nonetheless, to the best of our knowledge, video-based hallucinations remain underexplored in the existing literature.

\begin{wrapfigure}{r}{0.5\textwidth}
  \centering
    \caption{Multiple-Choice Question Answering performance of representative VLLMs on our \textsc{VidHal} benchmark. 
    (Left) Ranking of representative evaluated VLLMs. (Right) Detailed accuracy results for each temporal aspect, where higher scores indicate fewer hallucinations.}
    \includegraphics[width=\linewidth]{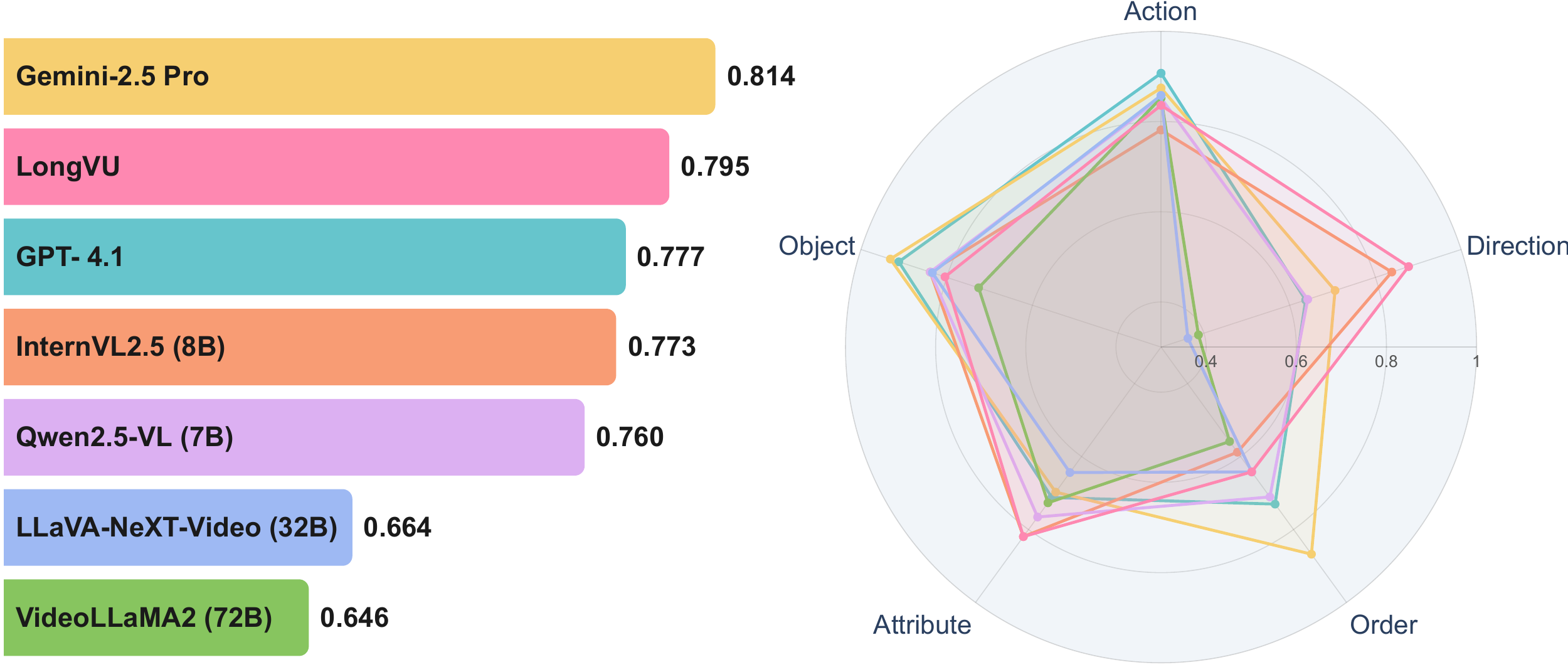}
    \label{fig:aspect_ranking_naive}
    \vspace{-1.25em}
\end{wrapfigure}
To address this research gap, we present \textsc{VidHal}, a benchmark specifically designed to evaluate video-based hallucinations of VLLMs. \textsc{VidHal} features videos that comprehensively cover a broad range of temporal aspects, such as entity actions and sequence of events. Each video is automatically annotated with multiple captions exhibiting \textit{varying levels} of aspect-specific hallucinations, capturing both subtle and significant discrepancies. In addition, we perform detailed human validation to ensure the robustness and reliability of our annotation process. An additional motivation stems from the limited metrics for quantifying hallucinations in VLLMs. 
To capture fine-grained hallucinatory errors of these models, we propose a unique caption ordering task that requires models to rank captions by hallucination levels, consequently leading to a novel ranking-based metric and a multiple-choice question answering metric, both distinct from prior binary-accuracy based studies and specifically tailored to evaluate nuanced hallucinations in video-based VLLMs.

Using our \textsc{VidHal} dataset, we benchmark thirteen VLLMs including both open-sourced and proprietary models, with abstracted results summarized in Figure \ref{fig:aspect_ranking_naive}. 
Through these extensive experiments, we identify limitations in nuanced video understanding among all evaluated VLLMs. 
Specifically, our findings reveal that existing VLLMs struggle to differentiate between captions with varying levels of hallucination.
This deficiency is particularly evident when evaluating video-specific aspects, such as \emph{Direction} and \emph{Order}, as illustrated in Figure~\ref{fig:aspect_ranking_naive}, indicating substantial room for improvement in current video-based VLLMs.
% Additionally, proprietary models, \eg, GPT-4o~\citep{openai2024gpt4technicalreport}, often outperform open-source counterparts by significant margins.

The contributions of this work are three-fold:
\begin{itemize}[leftmargin=10pt]
    \item We present \textsc{VidHal}, a benchmark dataset dedicated to video-based hallucination evaluation of VLLMs.
    Our dataset is distinguished by i) video instances encompassing a diverse range of temporal concepts and ii) captions with varying hallucination levels.
    \item We introduce a novel evaluation task of caption ordering along with two metrics designed to evaluate fine-grained hallucination generation in existing VLLMs.
    \item We conduct extensive experiments on \textsc{VidHal} with a variety of VLLMs, uncovering limitations in their fine-grained video reasoning abilities, particularly in their tendency to generate hallucinations.
\end{itemize}

\section{Related Work}

\noindent\textbf{Vision Large Language Models.}
The emergence of powerful LLMs has advanced the development of VLLMs. Typical methods in this category include LLaVA~\citep{liu2023visualinstructiontuning}, MiniGPT-4~\citep{zhu2023minigpt4enhancingvisionlanguageunderstanding}, InstructBLIP~\citep{dai2023instructblipgeneralpurposevisionlanguagemodels}, and Qwen-VL~\citep{2024pengqwen2vltechnical, 2025shuaiqwen25vltechnical}. These VLLMs rely on aligning vision encoders with LLMs using connective modules such as Q-Former~\citep{dai2023instructblipgeneralpurposevisionlanguagemodels, zhang2023videollamainstructiontunedaudiovisuallanguage, cheng2024videollama2advancingspatialtemporal} or MLPs~\citep{liu2024improvedbaselinesvisualinstruction, su2023pandagptmodelinstructionfollow} with the instruction tuning stage. Recent methods have extended visual inputs from images to (long) videos, delivering impressive joint spatial-temporal reasoning capabilities. For instance, VideoLLaMA2~\citep{cheng2024videollama2advancingspatialtemporal} enhances the LLaMA model with video understanding capabilities through a Spatial-Temporal Convolution (STC) module. LLaVA-NeXT-Video~\citep{liu2024llavanext, zhang2024llavanextvideo} presents an AnyRes approach that enables reasoning with long videos.

\noindent\textbf{Hallucinations in VLLMs.} Despite their impressive performance on visual reasoning benchmarks, current VLLMs remain notoriously susceptible to hallucinations~\citep{hacl, liu2024paying, zhu2024combating, unifiedhallucinationdetection}. A common demonstration is that generated responses contain information inconsistent with the visual content~\citep{liu2024surveyhallucinationlargevisionlanguage, fan2024helpd, xing2024efuf}. Most approaches address the hallucination problem with post-hoc techniques. For example, LURE~\citep{zhou2023analyzing} and Woodpecker~\citep{yin2023woodpecker} develop pipelines that assist VLLMs in revising their responses using expert models. To reduce bias from unimodal and statistical priors, contrastive decoding methods, such as VCD~\citep{leng2023mitigatingobjecthallucinationslarge} and M3ID~\citep{m3id}, along with attention calibration techniques like OPERA~\citep{huang2024operaalleviatinghallucinationmultimodal} are employed to refine token predictions. Building on the success of reinforcement learning in LLM development~\citep{ouyang2022traininglanguagemodelsfollow}, HA-DPO~\citep{zhao2024hallucinationsenhancinglvlmshallucinationaware}, POVID~\citep{povid} and CSR~\citep{csr} adopt this paradigm to fine-tune VLLMs, yielding outputs with fewer hallucinations.

\noindent\textbf{Video Reasoning Benchmarks.}
The rise of video-based VLLMs has driven the development of numerous video benchmarks. Notable examples, such as SEEDBench~\citep{li2023seedbenchbenchmarkingmultimodalllms}, VideoBench~\citep{ning2023videobenchcomprehensivebenchmarktoolkit}, MVBench~\citep{li2024mvbenchcomprehensivemultimodalvideo}, and VideoMME~\citep{fu2024videommefirstevercomprehensiveevaluation}, focus on dynamic events requiring temporal reasoning beyond individual frames. However, these benchmarks often lack diversity in reasoning tasks and visual concepts. To address this, AutoEval-Video~\citep{chen2024autoevalvideoautomaticbenchmarkassessing} and Perception Test~\citep{pătrăucean2023perceptiontestdiagnosticbenchmark} introduce complex reasoning tasks such as counterfactual and explanatory reasoning, while TempCompass~\citep{liu2024tempcompassvideollmsreally} expands temporal concept coverage.
Several benchmarks~\citep{li2023evaluatingobjecthallucinationlarge, wang2024amberllmfreemultidimensionalbenchmark, sun2024aligningmmhalbench, kaul2024throneobjectbasedhallucinationbenchmark, liu2024mitigatinghallucinationlargemultimodal, weitoward, multiobjecthallucination} have been constructed to quantify visual hallucinations, primarily targeting object-based hallucinations in images. HallusionBench~\citep{guan2024hallusionbenchadvanceddiagnosticsuite}, VideoCon~\citep{bansal2023videoconrobustvideolanguagealignment}, and Vript~\citep{vript} provide partial coverage of video-based hallucinations, while VidHalluc~\citep{vidhallucli2025} and VideoHallucer~\citep{videohallucer} introduce benchmarks for hallucination detection in videos. However, these benchmarks provide limited coverage of spatio-temporal concepts, focusing on conventional aspects like actions while neglecting other video-centric elements such as direction. \textit{Additionally, their evaluation strategies primarily follow image-based approaches, which we argue are less effective in capturing nuanced, video-specific hallucinations.}

\section{\textsc{VidHal} Dataset Construction}
\begin{figure*}[!t]
  \centering
  \footnotesize
    \caption{
   Overview of our \textsc{VidHal} benchmark construction pipeline. Using \emph{direction} as an example from the five selected aspects, we begin by sourcing relevant video instances from existing datasets. Next, the anchor (positive) caption is generated from the original video metadata. Finally, GPT-4o is employed to generate hallucinatory captions at varying levels.
   }
   % \includesvg[inkscapelatex=false, width=\linewidth]{dataset_overview}
   \includegraphics[width=\linewidth]{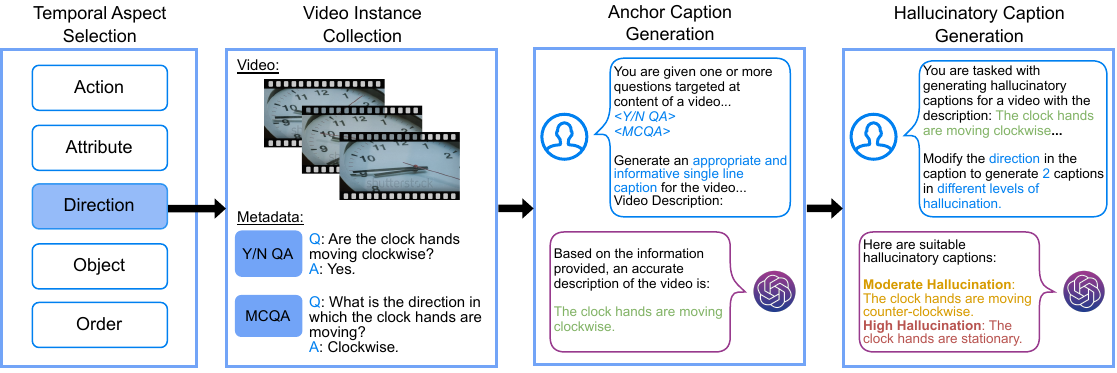}
   % \vspace{-1em}
   \label{fig:dataset_overview}
\end{figure*}

We introduce \textsc{VidHal}, a unique video-language benchmark designed to evaluate hallucinations of VLLMs in a comprehensive manner. As depicted in Figure \ref{fig:dataset_overview}, \textsc{VidHal} comprises of video instances which span a diverse spectrum of temporal aspects, including previously unexplored aspects such as directional movement. In contrast to previous studies on video hallucination evaluation~\citep{vript, videohallucer, vidhallucli2025}, \textsc{VidHal} incorporates multiple hallucinated captions per video, enabling the assessment of video hallucinations at multiple levels of granularity.

\subsection{Temporal Hallucinations in Videos}
Despite extensive studies on hallucinations in LLMs and VLLMs, the existing literature has not converged on a consensus formal definition. Following the majority of prior work in hallucination evaluation for VLLMs~\citep{li2023evaluatingobjecthallucinationlarge, videohallucer, vidhallucli2025}, we treat hallucinations as a specific class of errors in which the model generates content that directly contradicts visually grounded evidence. This is distinct from errors arising from imperfect reasoning or contextual analysis that do not necessarily conflict with the visual content. Compared to images, video hallucinations extend beyond static visual elements to include misperceptions of dynamic changes within scenes. We categorize these temporal hallucinations into two semantic levels:

\noindent\textbf{Lexical Semantics (L-Sem)} captures instances where VLLMs misinterpret words related to temporal features, including nouns referring to objects or attributes (e.g., misidentifying a color change from green to red as green to orange) and verbs describing actions (e.g., interpreting ``kicking a ball'' as ``throwing a ball'').

\noindent\textbf{Clause Semantics (C-Sem)} encompasses errors involving event descriptions and their sequences, where the VLLM incorrectly predicts the order of events occurring in the video. For example, given sequentially occurring events $A$ and $B$ in a video, the model may perceive $B$ preceding $A$.

By addressing these two dimensions of video-based hallucinations, \textsc{VidHal} offers holistic coverage over the level of detail in which VLLMs may hallucinate. This two-level decomposition is grounded in the temporal ontology ~\citep{moens1988temporal}, which formally establishes that both lexical and clause-level representations are necessary and sufficient for complete natural-language temporal description, spanning intra-event structure (L-Sem) through to inter-event relational ordering (C-Sem).

\subsection{Temporal Concept Selection}\label{subsec:concept_selection}
Prior research on hallucination evaluation for both images~\citep{li2023evaluatingobjecthallucinationlarge, wang2024amberllmfreemultidimensionalbenchmark,rohrbach2019objecthallucinationimagecaptioning} and videos~\citep{videohallucer,vript,guan2024hallusionbenchadvanceddiagnosticsuite} has predominantly focused on common visual aspects such as action- and object-based hallucinations. However, video-based hallucinations may involve additional dynamic factors associated with spatio-temporal patterns, which these studies overlook. In light of this, we propose to focus on the following five aspects to ensure comprehensive coverage of temporal concepts.
Specifically, the first four aspects address hallucinations based on lexical semantics, while the fifth targets clause semantics.

\begin{itemize}[leftmargin=10pt]
\setlength\itemsep{0.5em}
     \item \textbf{Attribute (L-Sem)} describes the fine-grained characteristics of objects or subjects in the video. We additionally categorize this aspect into sub-aspects of \textit{Size}, \textit{Shape}, \textit{Color}, \textit{Count} and \textit{State Change}. 
     \item \textbf{Object (L-Sem)} relates to the interactions between objects and entities within the video. We further delineate this aspect into two fine-grained sub-aspects: \textit{Object Recognition}, identifying the objects engaged in interactions, and \textit{Interaction Classification} which concentrate on how these objects interact with other objects or subjects.
    \item \textbf{Action (L-Sem)} refers to the movements and behaviours exhibited by entities.
    \item \textbf{Direction (L-Sem)} indicates the orientation and movement trajectory of subjects or objects.
    \item \textbf{Event Order (C-Sem)} represents the correct sequence of events in the video. During our collection, we retain videos that contain at least three distinct events.
\end{itemize}

The choice of these aspects is motivated by their collective corroboration across independent video-based studies~\citep{videohallucer, liu2024tempcompassvideollmsreally, li2024mvbenchcomprehensivemultimodalvideo}, which identify them as critical dimensions for robust spatio-temporal reasoning in VLLMs. Crucially, while certain aspects are not necessarily specific to videos (\eg, Object and Attribute), \textsc{VidHal} extends them beyond static spatial settings. For instance, attributes are evaluated on how they change dynamically over time rather than assessing them as fixed visual properties. We present an example that illustrates the direction aspect in Figure \ref{fig:dataset_overview}, with additional examples available in the supplementary material.

\subsection{Hallucinatory Caption Generation}\label{subsec:annotation_pipeline}

Based on the aspects defined in Section \ref{subsec:concept_selection}, we construct our benchmark from four public video understanding datasets: TempCompass~\citep{liu2024tempcompassvideollmsreally}, Perception Test~\citep{pătrăucean2023perceptiontestdiagnosticbenchmark}, MVBench~\citep{li2024mvbenchcomprehensivemultimodalvideo}, and AutoEval-Video~\citep{chen2024autoevalvideoautomaticbenchmarkassessing}. TempCompass and MVBench provide extensive coverage of all five temporal aspects, while Perception Test and AutoEval-Video focus on human-object interactions and attribute changes, respectively.

Existing hallucination benchmarks~\citep{li2023evaluatingobjecthallucinationlarge, wang2024amberllmfreemultidimensionalbenchmark} rely mostly on binary questions for evaluation, limiting their efficacy in detecting subtle video hallucinations, such as minor event inconsistencies. To address this issue, we advocate a novel evaluation protocol incorporating several carefully annotated captions. Specifically, each video will be annotated with $M$ captions that reflect varying degrees of hallucination in VLLMs. Given the cost and labor intensity of manual annotation, we follow existing benchmark studies such as PhD~\citep{liu2024phdpromptedvisualhallucination} and MVBench~\citep{li2024mvbenchcomprehensivemultimodalvideo}, opting for automatic caption generation using a carefully designed pipeline illustrated in Figure \ref{fig:dataset_overview}.

\begin{table*}[!t]
    
  \setlength{\tabcolsep}{2pt}
  \centering
    \caption{Comparison of our benchmark dataset with existing video-based reasoning and hallucination evaluation datasets. For datasets with multiple evaluation tasks, only those relevant to hallucination evaluation are included. VL Entailment denotes the task of \textit{video-language entailment}, while \textit{Event Ordering} prompts the model to determine the chronological sequence of scenes in a video.
  }
  \resizebox{1.0\linewidth}{!}{\begin{tabular}{@{}cccccccccccccc@{}}
    \toprule
    & \multirow{3}{*}{Dataset} & \multicolumn{10}{c}{Temporal Aspects} & \multirow{3}{*}{\makecell{Task\\Formats}} & \multirow{3}{*}{\makecell{Evaluation\\Metrics}} \\
        \cmidrule(lr){3-12}
        & & Action & \multicolumn{5}{c}{Attribute} & Direction & \multicolumn{2}{c}{Object} & Order & & \\
        \cmidrule(lr){4-8} \cmidrule(lr){10-11}
        & & & Size & Shape & Color & Count & State-Change & & Recognition & Interaction & & & \\
        \midrule
    % \multicolumn{1}{@{}l}{\textit{Video Understanding Benchmarks}} & & & \\
    \multirow[c]{4}{*}{\rotatebox[origin=c]{90}{\clap{%
        \parbox{2.25cm}{\centering \textit{Video\\Reasoning}}}}} 
    & SEEDBench~\citep{li2023seedbenchbenchmarkingmultimodalllms} & \cmark & \xmark & \xmark & \xmark & \xmark & \xmark & \xmark & \xmark & \xmark & \cmark & MCQA & Accuracy \\
    & VideoBench~\citep{ning2023videobenchcomprehensivebenchmarktoolkit} & \cmark & \cmark & \cmark & \cmark & \cmark & \xmark & \xmark & \cmark & \xmark & \xmark & MCQA & Accuracy\\
    & MVBench~\citep{li2024mvbenchcomprehensivemultimodalvideo} & \cmark & \xmark & \xmark & \xmark & \xmark & \xmark & \cmark & \cmark & \cmark & \cmark & MCQA & Accuracy\\
    & Video-MME~\citep{fu2024videommefirstevercomprehensiveevaluation} & \cmark & \cmark & \cmark & \cmark & \cmark & \xmark & \xmark & \cmark & \xmark & \xmark & MCQA & Accuracy\\
    \midrule
    % % \multicolumn{1}{@{}l}{\textit{Hallucination Evaluation Datasets}} & & & \\
    \multirow[c]{6}{*}{\rotatebox[origin=c]{90}{\clap{%
        \parbox{2cm}{\centering \textit{Hallucination\\Evaluation}}}}}
    & \multirow[c]{2}{*}{Vript~\citep{vript}} & \multirow[c]{2}{*}{\cmark} & \multirow[c]{2}{*}{\xmark} & \multirow[c]{2}{*}{\xmark} & \multirow[c]{2}{*}{\xmark} & \multirow[c]{2}{*}{\xmark} & \multirow[c]{2}{*}{\xmark} & \multirow[c]{2}{*}{\xmark} & \multirow[c]{2}{*}{\cmark} & \multirow[c]{2}{*}{\cmark} & \multirow[c]{2}{*}{\cmark} & Video Captioning & F1 Score \\
    & & & & & & & & & & & & Event Ordering & Accuracy \\[1pt]
    & VideoCon~\citep{bansal2023videoconrobustvideolanguagealignment} & \cmark & \cmark & \cmark & \cmark & \cmark & \xmark & \xmark & \cmark & \xmark & \cmark & VL Entailment & ROC-AUC \\
    & HallusionBench~\citep{guan2024hallusionbenchadvanceddiagnosticsuite} & \cmark & \xmark & \xmark & \xmark & \xmark & \xmark & \cmark & \xmark & \xmark & \cmark & Y/N QA & Accuracy \\[1pt]
    & \multirow[c]{2}{*}{\colorbox{green!20}{\textsc{VidHal} (Ours)}} & \multirow[c]{2}{*}{\cmark} & \multirow[c]{2}{*}{\cmark}  & \multirow[c]{2}{*}{\cmark}  & \multirow[c]{2}{*}{\cmark}  & \multirow[c]{2}{*}{\cmark}  & \multirow[c]{2}{*}{\cmark} & \multirow[c]{2}{*}{\cmark} & \multirow[c]{2}{*}{\cmark} & \multirow[c]{2}{*}{\cmark} & \multirow[c]{2}{*}{\cmark} & MCQA, & Accuracy \\
    & & & & & & & & & & & & Caption Ordering & NDCG \\
    \bottomrule
  \end{tabular}}
  % }
  % \vspace{-1.5em}
  \label{tab:dataset_comparison}
\end{table*}
\noindent \textbf{Anchor Caption Generation.} 
The video instances in \textsc{VidHal} are sourced from various public datasets, resulting in distinct associated metadata such as long-form captions in AutoEval-Video and question-answer pairs in MVBench. To ensure structure consistency and information granularity in the respective dataset description across all instances, we automatically generate an anchor caption for each video. Specifically, we input the metadata for each video $V^i$ into GPT-4o and prompt it to generate a concise and accurate description $y^i_+$ using the provided metadata information.

\noindent \textbf{Hallucinatory Caption Generation.} After obtaining the positive caption for each video instance, we augment the dataset with $M-1$ additional captions containing hallucinated content. For a given video instance $V^i$, we construct a set $\mathcal{Y}^i_- = \{ y^{i,1}_-, \cdots, y^{i, M-1}_-\}$ containing captions with different levels of hallucination based on the temporal concepts associated with it. Specifically, $y^{i,k}_-$ exhibits heavier hallucination than $y^{i,j}_-$ for caption hallucination degree $j < k$. We leverage GPT-4o to generate $\mathcal{Y}^i_-$ by combining the anchor caption $y^i_+$ and prompting it to create $y^{i,1}_-, \cdots, y^{i, M-1}_-$ progressively in increasing levels of hallucination, where the prompt includes detailed guidelines and definitions for the different grades of hallucination severity. Given the distinct characteristics of each aspect, hallucination severity is operationalized in terms of the likelihood of confusing the ground-truth aspect value with a plausible alternative, with aspect-specific criteria detailed in Table~\ref{tab:severity_criteria} in the Appendix. To steer GPT-4o towards generating captions that faithfully reflect the intended hallucination severity levels, the prompt further includes aspect-specific in-context examples that adhere to these operationalized criteria. The set of captions associated with $V^i$ is then defined as $\mathcal{Y}^i \leftarrow \{y^i_+\} \bigcup \mathcal{Y}^i_-$ consisting of both the anchor and hallucinatory captions. The full anchor and hallucinatory caption generation prompts, along with aspect-specific in-context examples, are detailed in Figures~\ref{fig:hallucination_caption_generation}--\ref{fig:in_context_action_direction} in the Appendix.

\subsection{Dataset Statistics and Human Validation}
% \begin{wrapfigure}{r}{0.5\textwidth}

% Our \textsc{VidHal} benchmark consists of a total of 1,000 video instances. 
Using our automatic annotation pipeline, our \textsc{VidHal} benchmark consists of a total of 1,000 video instances each tagged with $M\mathbin{=}3$ captions. As shown in Table \ref{tab:dataset_comparison}, our \textsc{VidHal} dataset stands out from other video understanding~\citep{li2023seedbenchbenchmarkingmultimodalllms, ning2023videobenchcomprehensivebenchmarktoolkit, li2024mvbenchcomprehensivemultimodalvideo, fu2024videommefirstevercomprehensiveevaluation} and hallucination benchmarks~\citep{guan2024hallusionbenchadvanceddiagnosticsuite, factvc} in terms of two dimensions:
\begin{enumerate*}[label=\Roman*)]
    \item \textsc{VidHal} encompasses a diverse range of video-centric temporal aspects; and
    \item We introduce a novel caption ordering task along with two tailored metrics to capture subtle hallucinations previously ignored by paired questions
\end{enumerate*}. 

\begin{figure}[!tb]
    \centering
     \caption{Human agreement on hallucination levels in the \textsc{VidHal} dataset. (Left) Distribution of agreement ratios per video sample. (Right) Average agreement ratio for each aspect, with an overall average of 87\%. The \textit{``Complete Agreement''} dotted line at 1.0 denotes the upper bound where the constructed caption orderings are consistent with human consensus rankings across all validated instances.}
    \begin{subfigure}[t]{0.49\linewidth}
         \centering
         \includegraphics[width=0.75\linewidth]{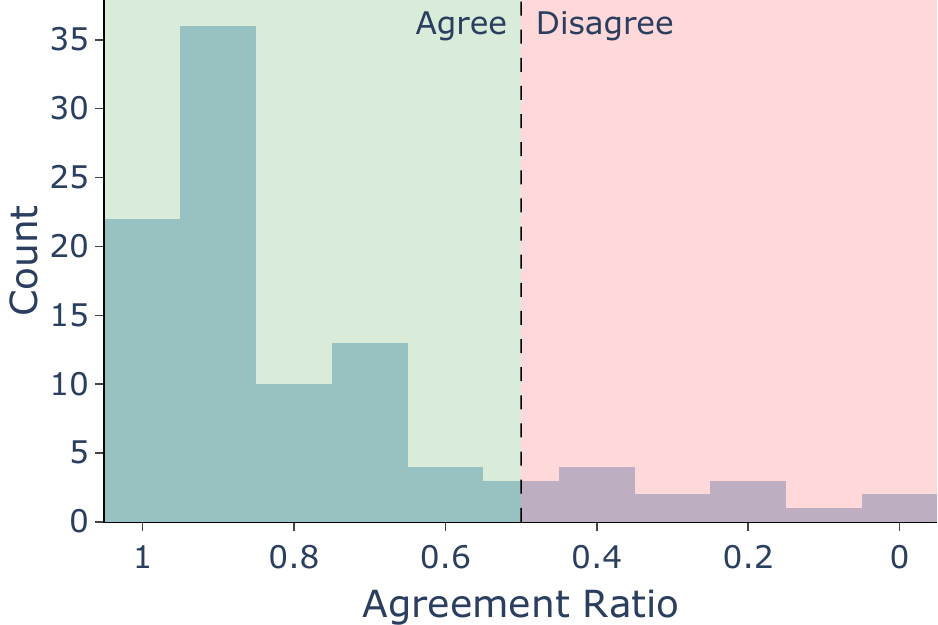}
         % \caption{A}
         % \label{fig:agreement_distribution}
     \end{subfigure}
     \hfill
     \begin{subfigure}[t]{0.49\linewidth}
         \centering
         \includegraphics[width=0.75\linewidth]{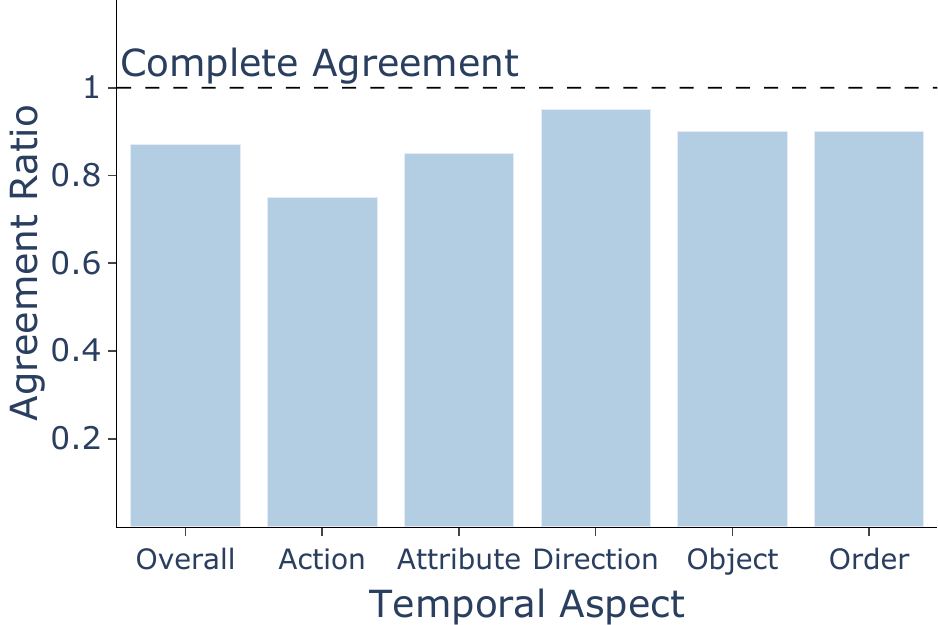}
         % \caption{B}
         % \label{fig:aspect_agreement}
     \end{subfigure}
     % \captionsetup{font=footnotesize}
     \label{fig:human_agreement}
\end{figure}

To validate the reliability of our generated captions, we randomly sampled 100 examples for human evaluation, with each example assessed by an average of 15 annotators. The validation process focused on verifying whether the ranking of hallucinatory captions produced by our pipeline aligns with human judgment. Specifically, the Inter-Rater agreement rate is computed as the proportion of validated instances in which our automatically generated caption ordering matches the consensus ranking derived from human annotations, where the consensus is determined by a majority vote among individual annotator rankings. As shown in Figure \ref{fig:human_agreement}, an overall agreement rate of 87\% was achieved, demonstrating strong consistency with human preferences across all temporal aspects.
\section{\textsc{VidHal} Evaluation Protocol}\label{sec:evaluation_protocol}

To address the limitations of binary question-based benchmarks, we propose two evaluation tasks: \textit{multiple-choice question answering} and a novel \textit{caption ordering} task, detailed in Section \ref{subsec:evaluation_tasks}. Additionally, we develop corresponding metrics to comprehensively measure hallucinations in video-based VLLMs in Section \ref{subset:evaluation_metrics}.

\subsection{Evaluation Tasks}\label{subsec:evaluation_tasks}
\noindent \textbf{Multiple-Choice Question Answering (MCQA)} assesses the model's spatiotemporal understanding in a coarse-grained manner. Specifically, the VLLM is provided with a video $V^i$ and its corresponding set of captions $\mathcal{Y}^i$ as answer options and instructed to select the most appropriate caption for the video.\\

\noindent \textbf{Caption Ordering} evaluates a model’s visual reasoning from a nuanced granularity, instructing VLLMs to order the provided captions based on their hallucination level. Through pairwise comparisons across all captions, this task identifies cases where the model struggles to distinguish varying levels of hallucination severity beyond anchor-hallucination distinctions.

Specifically, we design two caption ordering sub-tasks. The first, \textit{naive caption ordering}, requires VLLMs to rank all captions at once. However, this sub-task can confuse several VLLMs due to its inherently challenging nature and the inferior instruction-following capabilities of some models. As a complement, we propose an additional sub-task, \textit{relative caption ordering}, which decomposes the prior task into multiple paired caption ordering tasks. Since each paired ordering task is answered in isolation, the VLLM may produce a non-transitive, cyclic ranking. 

\begin{wrapfigure}{r}{0.5\textwidth}
  \centering
  \vspace{-1.25em}
  \caption{Visual illustration of the \textit{relative caption ordering} task in \textsc{VidHal}.}
   \includegraphics[width=0.9\linewidth]{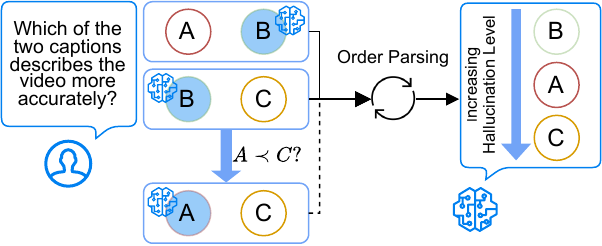}
   \label{fig:evaluation_protocol}
   % \vspace{-0.75em}
\end{wrapfigure}

To circumvent this, we query the model with consecutive caption pairs, prompting the final pair only if multiple orderings are possible. For instance, given captions $A$, $B$, and $C$, if the model predicts $A \prec B$ and $B \prec C$, the overall order $A \prec B \prec C$ can be directly inferred. However, if it instead ranks $B \prec A$ , as shown in Figure~\ref{fig:evaluation_protocol}, we additionally include a third comparison between $A$ and $C$ to resolve any ambiguity in determining in the final order.
Notably, our relative caption ordering task is more challenging than previous binary questions. This complexity arises from certain paired questions in \textsc{VidHal} where both options are hallucinatory, making them harder to distinguish as opposed to \textlangle\textit{positive, hallucinatory}\textrangle\xspace pairs.

\subsection{Evaluation Metrics}\label{subset:evaluation_metrics}
\paragraph{Notations} For a particular video instance $V^i$, we define the ground truth caption order for $V^i$ to be $\mathcal{Y}^i_* = (y^i_+, y^{i,1}_-, \cdots, y^{i, M-1}_-)$. Further let the $j^{th}$ element in this ordering be indexed as $\mathcal{Y}^{i, j}_*$.
\paragraph{MCQA} We employ the standard accuracy metric for the MCQA task:
\begin{equation}
    \text{Accuracy} = \frac{1}{N} \sum^N_{i=1} \mathbb{I}\left[R_{MCQA}(V^i, \mathcal{Y}^i) = y^i_+\right],
\end{equation}
where $N$ is the number of video instances, $\mathbb{I}$ denotes the indicator function, and $R_{MCQA}(V^i, \mathcal{Y}^i)$ represents the best matched caption from $\mathcal{Y}^i$ for $V^i$ as predicted by a VLLM. 
\paragraph{Caption Ranking} 
Inspired by metrics from the information retrieval domain~\citep{chat-rec}, we adapt the well-established Normalized Discounted Cumulative Gain (NDCG)~\citep{2002ndcg} for hallucination assessment in \textsc{VidHal}. Unlike previous metrics like POPE~\citep{li2023evaluatingobjecthallucinationlarge}, our metric awards partial credit for correctly ordered caption pairs even when the optimal ranking is not achieved. As such, we expect the metric to effectively capture and distinguish both subtle and severe hallucinations generated by video-based VLLMs. Formally, we define our adapted NDCG metric as follows:
\begin{equation}
    \text{NDCG} = \frac{1}{N}\sum^N_{i=1} \frac{\text{DCG}_i - \text{rDCG}_i}{\text{iDCG}_i - \text{rDCG}_i},
\end{equation}
where $\text{DCG}_i$ is formulated as:
\begin{equation}
    \text{DCG}_i = \sum^M_{j=1} \frac{r\left( \hat{y}^{i,j}, \mathcal{Y}^i_* \right)}{\log (j + 1)},
\end{equation}
and $\hat{y}^{i,j}$ represents $j^{th}$ caption in the ranked order predicted by the VLLM. The perfect ordering is achieved when $\hat{y}^{i,1}=y^i_+$ and $\{\hat{y}^{i, j} = y^{i, j-1}_{-}\}_{j=2 \to M}$. To evaluate predicted caption orders relative to this ideal sequence, a relevance function $r\left(\hat{y}^{i,j}, \mathcal{Y}^i_* \right)$ is designed to assign higher scores to $\hat{y}^{i,j}$ with lower hallucinatory extent: 
% We formalize the relevance function below:
\begin{equation}
    r(\hat{y}^{i,j}, \mathcal{Y}^i_*) = M + 1 - \text{pos}(\hat{y}^{i,j}, \mathcal{Y}^i_*),
\end{equation}
where $\text{pos}(\hat{y}^{i,j}, \mathcal{Y}^i_*)$ denotes the position of $\hat{y}^{i,j}$ in $\mathcal{Y}^i_*$. Finally, $\text{DCG}_i$ is normalized to a range of $[0, 1]$ using $\text{iDCG}_i$ and $\text{rDCG}_i$,
with a score of 1 indicating perfect alignment of the predicted order with $\mathcal{Y}^i_*$. Specifically, these terms represent the maximum and minimum $\text{DCG}_i$ scores obtained from the optimal ordering $\mathcal{Y}^i_*$ and its reverse, respectively,
\begin{equation}
    \text{iDCG}_i = \sum^M_{j=1} \frac{r\left(\mathcal{Y}^{i, j}_*, \mathcal{Y}^i_* \right)}{\log (j + 1)}, \
    \text{rDCG}_i = \sum^M_{j=1} \frac{r\left(\mathcal{Y}^{i, M - j}_*, \mathcal{Y}^i_* \right)}{\log (j + 1)}.
\end{equation}

\section{Experiments}
\subsection{Experimental Settings}
\noindent\textbf{Models.} We evaluated twenty-three VLLMs from thirteen different model families, including ten open-source models: VideoChat~\citep{videochat2023}, LLaMA-VID~\citep{llamavideccv2024}, VideoChat2~\citep{li2024mvbenchcomprehensivemultimodalvideo}, mPLUG-Owl3~\citep{ye2024mplugowl3longimagesequenceunderstanding}, LLaVA-NeXT-Video~\citep{zhang2024llavanextvideo}, VideoLLaMA2~\citep{cheng2024videollama2advancingspatialtemporal}, MiniCPM-V~\citep{yao2024minicpm}, LongVU~\citep{2024shenlongvu}, InternVL2.5~\citep{2024cheninternvl25} and Qwen2.5-VL~\citep{2025shuaiqwen25vltechnical}, and three proprietary models: GPT-4o~\citep{openai2024gpt4technicalreport}, GPT-4.1 and Gemini~\citep{geminiteam2024gemini15unlockingmultimodal, 2025gemini25google}. These models represent a wide variety of architectural designs and training paradigms. Additionally, we included a random baseline that selects and ranks candidate options randomly.

\noindent\textbf{Implementation Details.} All experiments were conducted using four NVIDIA A100 40GB GPUs and inference APIs. The input captions in $\mathcal{Y}^i$ were randomized using a fixed, predefined randomization seed across experiments.
We adhered to the inference and model hyperparameters outlined in the respective original models, and employed greedy decoding during generation for a fair comparison.

\begin{table*}[!tb]
  \centering
  \caption{Overall benchmark performance of VLLMs on our \textsc{VidHal} dataset. \#Params refers to the number of parameters of the base LLM used. The best performance for each task is highlighted in \textbf{bold} for open-source models, and \underline{underlined} for proprietary models.}
  \midruletight
  \resizebox{1.0\linewidth}{!}{
  \begin{tabular}{lrrrcccc}
    \toprule
    \multirow{2}{*}{Model} & \multirow{2}{*}{\makecell{Vision Encoder}} & \multirow{2}{*}{LLM} & \multirow{2}{*}{\#Params} & \multirow{2}{*}{\#Frames} & \multirow{2}{*}{Accuracy} & \multicolumn{2}{c}{NDCG} \\
    \cmidrule{7-8}
    & & & & & & Naive & Relative \\
    \midrule
    \multicolumn{1}{@{}l}{\textit{Baseline}} & & & & & & & \\
    \midrule 
    \multicolumn{1}{l|}{Random} & - & - & - & \multicolumn{1}{c|}{-} & 0.326 & 0.505 & 0.480 \\
    \midrule
    \multicolumn{1}{@{}l}{\textit{Open-Source Models}} & & & & & & & \\
    \midrule
    \multicolumn{1}{l|}{VideoChat} & EVA-CLIP-G & Vicuna & 7B & \multicolumn{1}{c|}{8} & 0.381 & 0.475 & 0.488 \\
    \multicolumn{1}{l|}{LLaMA-VID} & EVA-CLIP-G & Vicuna & 7B & \multicolumn{1}{c|}{1fps} & 0.358 & 0.486 & 0.521 \\
    \multicolumn{1}{l|}{VideoChat2 (Vicuna)} & UMT-L & Vicuna & 7B & \multicolumn{1}{c|}{16} & 0.426 & 0.486 & 0.577 \\
    \multicolumn{1}{l|}{VideoChat2 (Mistral)} & UMT-L & Mistral & 7B & \multicolumn{1}{c|}{16} & 0.443 & 0.503 & 0.475 \\
    \multicolumn{1}{l|}{VideoChat2 (Phi)} & UMT-L & Phi3 & 3.8B & \multicolumn{1}{c|}{16} & 0.514 & 0.626 & 0.612 \\
    \multicolumn{1}{l|}{mPLUG-Owl3} & SigLIP/SO400M & Qwen2 & 7B & \multicolumn{1}{c|}{16} & 0.596 & 0.641 & 0.707 \\
    \multicolumn{1}{l|}{LLaVA-NeXT-Video (7B)} & SigLIP/SO400M & Vicuna & 7B & \multicolumn{1}{c|}{32} & 0.509 & 0.518 & 0.620 \\
    \multicolumn{1}{l|}{LLaVA-NeXT-Video (32B)} & SigLIP/SO400M & Qwen1.5 & 32B & \multicolumn{1}{c|}{32} & 0.663 & 0.641 & 0.747 \\
    \multicolumn{1}{l|}{VideoLLaMA2 (7B)} & CLIP ViT-L/14 & Mistral & 7B & \multicolumn{1}{c|}{8} & 0.541 & 0.564 & 0.622 \\
    \multicolumn{1}{l|}{VideoLLaMA2 (72B)} & CLIP ViT-L/14 & Qwen2 & 72B & \multicolumn{1}{c|}{8} & 0.647 & 0.787 & 0.760 \\
    \multicolumn{1}{l|}{MiniCPM-V 2.6} & SigLIP/SO400M & Qwen2 & 7B & \multicolumn{1}{c|}{1fps} & 0.377 & 0.530 & 0.523 \\
    \multicolumn{1}{l|}{LongVU} & SigLIP/SO400M & Qwen2 & 7B & \multicolumn{1}{c|}{1fps} & \textbf{0.795} & 0.453 & \textbf{0.846} \\
    \multicolumn{1}{l|}{InternVL2.5 (8B)} & InternViT-300M (V2.5) & InternLM2.5 & 7B & \multicolumn{1}{c|}{16} & 0.773 & 0.475 & 0.827 \\
    \multicolumn{1}{l|}{InternVL2.5 (26B)} & InternViT-6B (V2.5) & InternLM2.5 & 20B & \multicolumn{1}{c|}{16} & 0.742 & 0.498 & 0.775 \\
    \multicolumn{1}{l|}{Qwen2.5-VL (7B)} & Qwen2.5-ViT & Qwen2.5 & 7B & \multicolumn{1}{c|}{1fps} & 0.76 & \textbf{0.825} & 0.826 \\
    \multicolumn{1}{l|}{Qwen2.5-VL (32B)} & Qwen2.5-ViT & Qwen2.5 & 32B & \multicolumn{1}{c|}{1fps} & 0.732 & 0.811 & 0.800 \\
    \multicolumn{1}{l|}{Qwen2.5-VL (72B)} & Qwen2.5-ViT & Qwen2.5 & 72B & \multicolumn{1}{c|}{1fps} & 0.74 & 0.807 & 0.793 \\
    \midrule
    \multicolumn{1}{@{}l}{\textit{Proprietary Models}} & & & & & & & \\
    \midrule
    \multicolumn{1}{l|}{GPT-4o} & - & - & - & \multicolumn{1}{c|}{1fps} & 0.772 & 0.840 & 0.826 \\
    \multicolumn{1}{l|}{GPT-4.1} & - & - & - & \multicolumn{1}{c|}{1fps} & 0.777 & 0.845 & 0.834 \\
    \multicolumn{1}{l|}{Gemini-1.5 (Flash)} & - & - & - & \multicolumn{1}{c|}{1fps} & 0.657 & 0.738 & 0.745\\
    \multicolumn{1}{l|}{Gemini-1.5 (Pro)} & - & - & - & \multicolumn{1}{c|}{1fps} & 0.671 & 0.765 & 0.753 \\
    \multicolumn{1}{l|}{Gemini-2.5 (Flash)} & - & - & - & \multicolumn{1}{c|}{1fps} & 0.814 & 0.875 & 0.860\\
    \multicolumn{1}{l|}{Gemini-2.5 (Pro)} & - & - & - & \multicolumn{1}{c|}{1fps} & \underline{0.814} & \underline{0.876} & \underline{0.861} \\
    \bottomrule
  \end{tabular}}
  \label{tab:results}
  % \vspace{-1.5em}
\end{table*}
\subsection{Overall Results}

\paragraph{Benchmark Results.} We present the overall results of representative VLLMs in Table \ref{tab:results} across both MCQA and caption ordering tasks. We make three key observations from this table: 

\noindent\textit{Competitive Performance of Open-Source Models.} Open-source VLLMs achieve performance comparable to proprietary models, particularly on MCQA and relative caption ordering tasks. Notably, LongVU achieves the highest performance among open-source models and surpasses strong proprietary models such as GPT-4o, GPT-4.1, and Gemini-1.5 on these tasks.

\noindent\textit{Parameter Scale vs. Performance.} Among open-source VLLMs, smaller variants (e.g., 7B parameter models) outperform their larger counterparts within the same model family, as observed with InternVL2.5 and Qwen2.5-VL. This suggests that simply increasing model capacity may provide limited benefits for reducing video-based hallucinations in current VLLM development. While LLaVA-NeXT-Video and VideoLLaMA2 use different backbone LLMs in their larger variants, the performance gains observed are more likely attributable to increased parameter scale rather than the choice of LLM.

\noindent\textit{Impact of Architecture Design.} Model families that achieve high scores across both tasks often incorporate design efforts specifically targeting visual understanding, such as dynamic resolution scaling (InternVL2.5, Qwen2.5-VL) and temporal reduction techniques (LongVU). These findings may suggest that specialized architectural innovations are key factors in mitigating temporal hallucinations.

\begin{figure}[!tb]
    \centering
    \footnotesize
    \caption{Performance of VLLMs across individual aspects and sub-aspects in \textsc{VidHal}. (Left) Aspect-specific NDCG scores for the naive and relative caption ordering. (Middle) NDCG scores for \textit{Attribute}, and (Right) \textit{Object} sub-aspects in relative caption ordering.}
    \includegraphics[width=0.9\linewidth]{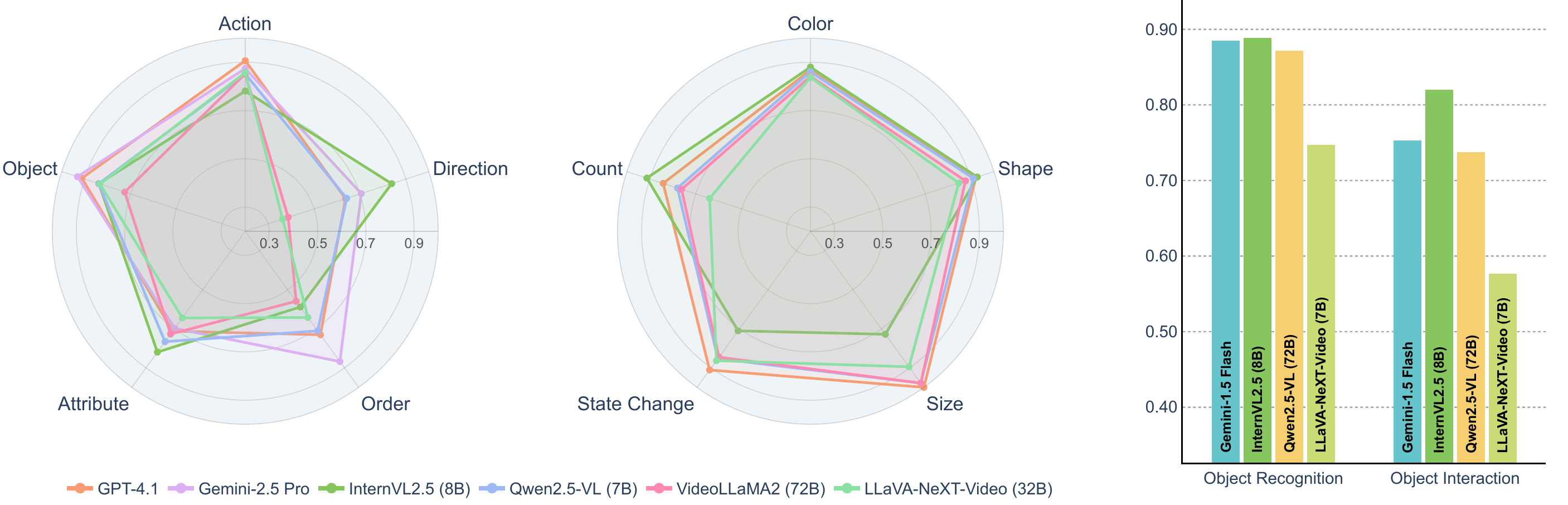}
    \label{fig:combined_performance}
    % \vspace{-1.5em}
\end{figure}

\paragraph{Aspect-aware Results.} Figure~\ref{fig:combined_performance} (Left) highlights the fine-grained, aspect-specific performance of the notable VLLMs. Notably, VLLMs demonstrate substantially stronger results on the \textit{Action} and \textit{Object} aspects compared to others. This can likely be attributed to current visual instruction tuning datasets predominantly emphasizing object-centric recognition and coarse-grained activity classification, potentially encouraging strong reliance on image-based priors when generating predictions. In contrast, these models tend to underperform on temporally nuanced aspects such as direction and event order, which are inherently unique to the video modality.

% Sub-aspect analysis
We further analyzed the distribution of results for the relative caption ranking task across sub-aspects of the \textit{Attribute} and \textit{Object} aspects in the middle and rightmost subfigures in Figure \ref{fig:combined_performance}. While VLLMs generally maintain consistent performance across \textit{Attribute} sub-aspects, their effectiveness declines slightly when reasoning about \textit{Count} and \textit{Color}, suggesting that reasoning over such fine-grained visual properties remains challenging for VLLMs. For the \textit{Object} aspect, several models performed significantly worse in \textit{Interaction Classification} than in \textit{Object Recognition}, highlighting the need to better model object interactions to bridge the gap between recognition and understanding.

\subsection{Ablation Studies}
We conducted further experiments to investigate VLLM behaviour beyond the aggregate benchmark results presented in Section~\ref{tab:results}, analysing model behaviour and failure modes in greater detail. Specifically, we begin by characterising \textit{where} models fail, examining pairwise error rates across caption pairs of varying hallucination levels to identify where nuanced discrimination breaks down. Next, we probe a plausible underlying reason for \textit{why} such failures arise, exploring the extent to which models over-rely on single-frame spatial information rather than leveraging full temporal context. Finally, we validate the robustness of our evaluation protocol against potential artifacts introduced by the relative caption ordering task.
 
\paragraph{Hallucination Differentiation Sensitivity.} To characterise \textit{where} models fail, we studied VLLM robustness in differentiating caption pairs of varying hallucination levels, beyond overall ordering performance.
% \begin{figure}[!b]
%     \centering
%     \caption{Hallucination misalignment (HM) scores on \textsc{VidHal}, with \textit{Random} representing HM scores from the random baseline.}
%     \includegraphics[width=0.75\linewidth]{figures/hallucination_misalignment.pdf}
%     \label{fig:hallucination_misalignment}
% \end{figure}
Specifically, we identify how the tendency of VLLMs to favour captions with higher hallucination over those with lower degree varies as the difference in hallucination severity between the compared captions changes. Formally, for two captions with different hallucination levels $j, k$ where $j > k$, we introduce the following metric to quantify such \emph{hallucination misalignment} cases:
 \begin{equation}
    HM_{j \rightarrow k} = \frac{1}{N} \sum^N_{i=1} \mathbb{I}\left[\mathcal{Y}^{i,j}_* \prec \mathcal{Y}^{i,k}_*\right],
\end{equation}
\begin{wrapfigure}{r}{0.55\linewidth}
    \centering
    \vspace{-1.25em}
    \caption{Hallucination misalignment (HM) scores on \textsc{VidHal}, with \textit{Random} representing HM scores from the random baseline.}
    \includegraphics[width=\linewidth]{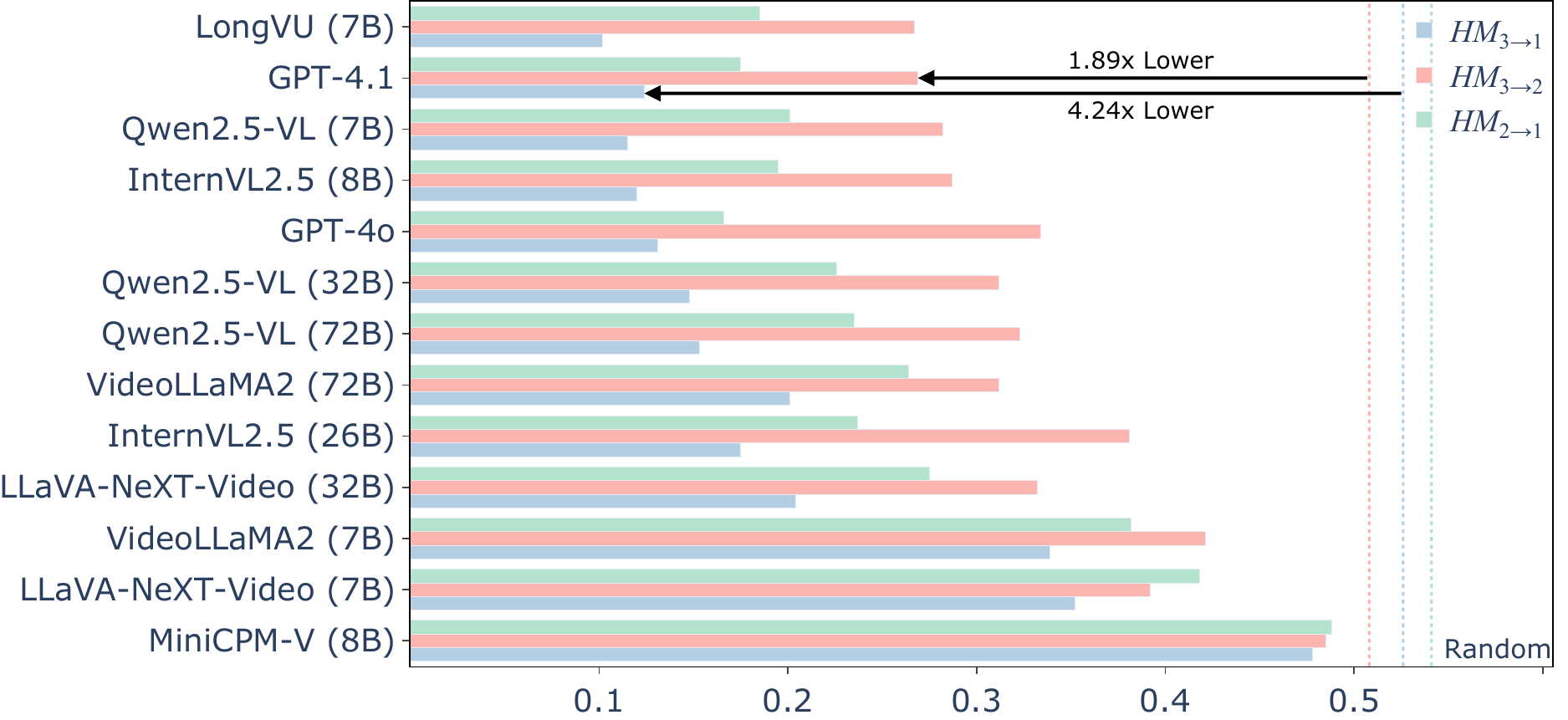}
    \label{fig:hallucination_misalignment}
    \vspace{-2em}
\end{wrapfigure}
which reflects the proportion of cases in which the VLLM selects the caption with a higher level of hallucination $j$ over $k$. Specifically, we examine three key cases: when the most hallucinatory caption is chosen over both the lower-hallucination and anchor captions, and when the lower-hallucination caption is selected over the anchor caption. These cases are represented by $HM_{3 \rightarrow 1}$, $HM_{3 \rightarrow 2}$, and $HM_{2 \rightarrow 1}$, respectively, with results presented in Figure \ref{fig:hallucination_misalignment}.

Our findings show that advanced VLLMs, such as VideoLLaMA2 (72B), GPT-4.1 and Qwen2.5-VL models can generally distinguish positive captions from severely hallucinated ones, reflected by their low $HM_{3 \rightarrow 1}$ scores in Figure \ref{fig:hallucination_misalignment}. However, two key observations emerge from our experiments: First, most VLLMs struggle to differentiate the lower hallucinatory caption from the anchor, as evidenced by the gap between $HM_{3 \rightarrow 1}$ and $HM_{2 \rightarrow 1}$. Second, all models exhibit high $HM_{3 \rightarrow 2}$ scores, indicating difficulty in distinguishing between two hallucinatory captions with varying degrees. These results suggest gaps in nuanced video reasoning may contribute to hallucinatory behavior in VLLMs, a challenge not addressed by existing \textlangle \textit{positive}, \textit{hallucinatory}\textrangle-based evaluation methods.~\citep{li2023evaluatingobjecthallucinationlarge, videohallucer, guan2024hallusionbenchadvanceddiagnosticsuite}.

\paragraph{Image Prior Reliance.} 
\begin{figure}[!t]
    \centering
    \caption{Overlapping ratios of model predictions under single-frame and full-video inputs for correct, incorrect and overall predictions in the (Left) naive and (Right) relative caption ordering tasks. \textit{Complete Reliance} indicates that the VLLM always produces the same response for both video and single frames.}
    \includegraphics[width=0.95\linewidth]{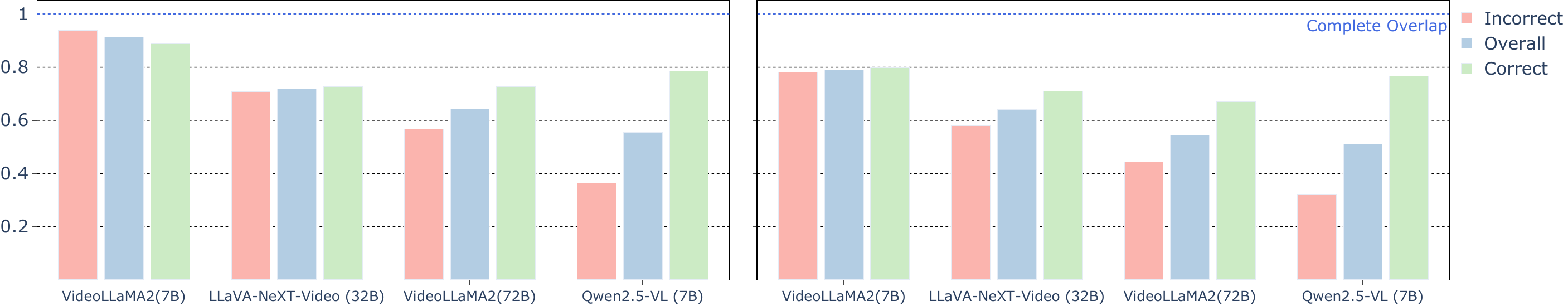}
    \label{fig:image_prior_reliance}
\end{figure}
Next, we investigated a plausible underlying cause for \textit{why} such deficiencies are observed, examining whether poor performance stems from models over-prioritising spatial over temporal information. Previous research shows that VLLMs often rely on image priors for reasoning~\citep{singleframebias2023, revisiting2022}, overlooking key spatiotemporal features. This is exemplified by dominant influence of a few frames on response generation. To examine how this bias affects video-based hallucinations, we used a video summarization algorithm~\citep{csta} to extract the most salient frame $v^i$ from $V^i$. We then generated VLLM responses on \textsc{VidHal} using $v^i$ instead of $V^i$ as visual input. The effect of image priors is evaluated by identifying overlapping instances where responses from $V^i$ and $v^i$ remain consistent across both correct and incorrect orderings. As shown in Figure \ref{fig:image_prior_reliance}, results reveal that VLLMs heavily rely on image priors, particularly in smaller models where overlap ratios are consistently high across correct, incorrect, and overall predictions alike. With increasing model capacity, the overlap ratios for incorrect predictions decrease, suggesting that larger models make better use of additional temporal information to correct errors, reflecting more robust spatio-temporal reasoning. Notably, Qwen2.5-VL (7B) demonstrates comparatively lower overlap ratios, particularly for incorrect predictions, relative to models of similar or even greater parameter scale, indicating a stronger capacity to leverage full-video temporal cues to refine its predictions.

\paragraph{Transitive Robustness.}
\begin{table}[!tb]
  \centering
  \caption{Proportion of \textsc{VidHal} instances where the predicted caption ordering $\mathcal{\hat{Y}}$ is cyclic.}
  \footnotesize
  \begin{tabular}{@{}lcccc@{}}
    \toprule
    Metric & VideoChat2 (7B) & mPLUG-Owl3 & LLaVA-NeXT-Video (7B) & VideoLLaMA2 (7B) \\
    \midrule
    $P(\text{IsCyclic}(\mathcal{\hat{Y}}))$ & 0.073 & 0.030 & 0.037 & 0.040 \\
    \bottomrule
  \end{tabular}
  \label{tab:transitivity_analysis}
\end{table}
Lastly, since the relative caption ordering task decomposes ordering into sequential pairwise comparisons, we verify that cyclic preferences do not confound the evaluation scores. Although the relative caption ranking task enhances VLLM stability in caption ordering, it may reveal transitive inconsistencies between the final caption order and individual paired orders. For instance, the predicted order $\mathcal{\hat{Y}}^i = [\hat{y}^{i,1}, \hat{y}^{i,2}, \hat{y}^{i,3}]$ may be inferred when the VLLM predicts $\hat{y}^{i,1} \prec \hat{y}^{i,2}$ and $\hat{y}^{i,2} \prec \hat{y}^{i,3}$, but it may predict $\hat{y}^{i,3} \prec \hat{y}^{i,1}$ when prompted to order these two captions, contradicting the initial order. We quantify such errors by measuring the proportion of instances where $\mathcal{\hat{Y}}^i$ fails to entail $\hat{y}^{i,3} \prec \hat{y}^{i,1}$. This is defined as:
\begin{equation}
P(\text{IsCyclic}(\mathcal{\hat{Y}})) = \frac{1}{N} \sum^N_{i=1} \mathbb{I}\left[R_{MCQA}(V^i, \{\hat{y}^{i,1}, \hat{y}^{i,3}\}) = \hat{y}^{i,3} \right].
\end{equation}
The results, presented in Table \ref{tab:transitivity_analysis}, demonstrate that even the least performant models exhibit strong consistency between pairwise caption preferences and final caption orderings. These results indicate that our proposed caption-ordering task is robust against anomalous scoring stemming from unintended cyclic preferences.

\subsection{Qualitative Results} 
\begin{figure*}[!b]
    \centering
    \caption{Qualitative examples of VLLM responses on the caption ordering tasks, for the \textit{Attribute}, \textit{Order} and \textit{Action} aspects. Captions are labelled $A$, $B$, and $C$ in increasing order of hallucination level, with the ground truth ordering $A \succ B \succ C$. Predicted orderings are shown for both naive and relative caption ordering across representative VLLMs.}
    \includegraphics[width=\linewidth]{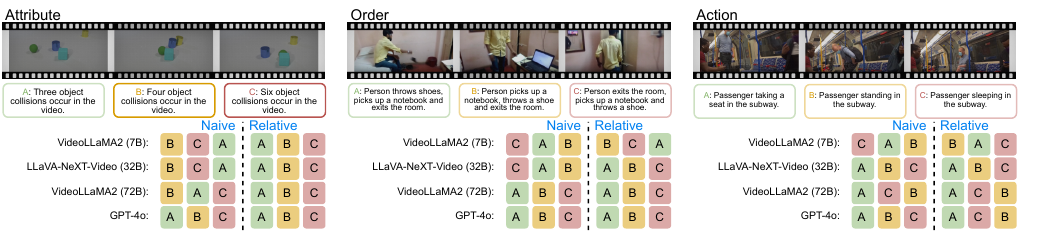}
    \label{fig:qualitative_examples}
\end{figure*}
We conducted a qualitative analysis of VLLM-generated responses for the caption ordering task to complement the quantitative results with deeper insight into model behaviour across task formats. We randomly sampled representative instances across the temporal aspects, evaluated selected VLLMs on both the naive and relative caption ordering tasks, and recorded their predicted orderings for direct comparison within each instance. The Action, Attribute, and Order aspects were selected for illustration in Figure~\ref{fig:qualitative_examples}. For each instance, the three captions are labelled A, B, and C in increasing order of hallucination level, with the ground truth ordering defined as $A \succ B \succ C$. We assess reasoning strength and consistency by examining the number of adjacent swaps required to transform each predicted ordering into the optimal one. Our analysis reveals two notable patterns: 

\noindent\textit{Relative caption ordering consistently elicits more accurate VLLM responses compared to naive ordering}, with this improvement being more pronounced in weaker models. The enhanced performance is evidenced by increased ordering correctness when transitioning from naive to relative ordering, as pairwise comparison prompts enable models to make finer-grained distinctions between caption quality levels.

\noindent\textit{Advanced VLLMs demonstrate greater stability across both ordering paradigms}, exhibiting lower variance in predictions between both ordering tasks. Stronger models such as VideoLLaMA2 (72B) and GPT-4o generally require at most one swap to reach the optimal ordering, consistent with their higher NDCG scores in Table~\ref{tab:results}. Furthermore, their predicted orderings under naive and relative caption ordering differ by at most one positional swap, as observed in the Attribute and Action aspects. This consistency suggests that stronger models maintain more coherent internal representations of caption quality across varying task formats.
\section{Conclusion}
\noindent\textbf{Summary.} In this work, we introduce the \textsc{VidHal} benchmark to address gaps in the video-based hallucination evaluation of VLLMs. \textsc{VidHal} features video instances spanning five temporal aspects. Additionally, we propose a novel caption ordering evaluation task to probe the fine-grained video understanding capabilities of VLLMs. We conduct extensive experiments on \textsc{VidHal} through the evaluation of twenty-three VLLMs, exposing their limitations in unexpected hallucination generation. 
Our empirical results shed light on several promising directions for future work: 
\eg, incorporating a broader range of temporal features during pretraining and mitigating single-frame priors to enhance temporal reasoning. 
These advancements will help to address the hallucination problem in video-based VLLMs, enhancing their robustness for real-world video understanding applications.

\noindent\textbf{Limitations.} We acknowledge that the \textsc{VidHal} evaluation suite relies on synthetic captions generated by GPT-4o, which may contain biases inherently present in the model. We note that this design choice is consistent with prior research, as several established language-only and vision-language benchmarks similarly use GPT-4o for dataset construction \citep{liu2024phdpromptedvisualhallucination, vidhallucli2025, li2024mvbenchcomprehensivemultimodalvideo, li2023seedbenchbenchmarkingmultimodalllms, haluevalli2023} or response evaluation \citep{guan2024hallusionbenchadvanceddiagnosticsuite, sun2024aligningmmhalbench, liu2024mitigatinghallucinationlargemultimodal}. To reduce over-alignment to GPT-4o's preferences, we incorporate additional strong LLMs, including Gemini-1.5 \citep{geminiteam2024gemini15unlockingmultimodal} and LLaMA2 (70B) \citep{llama2meta} to assess and filter generated captions. We further conduct a final step of manual verification and editing to address residual misalignments not captured by automated filtering. Full details of this post-processing pipeline are provided in Appendix~\ref{sec:dataset_dev}. While these measures enhance annotation robustness, fully eliminating LLM-induced biases in synthetic caption generation remains an open challenge. Moreover, while \textsc{VidHal} is designed to assess both subtle and severe hallucinations across a wide range of temporal aspects, the generated hallucinatory captions may not fully capture the internal tendencies of all evaluated VLLMs, a caveat researchers should bear in mind when interpreting results.

\textsc{VidHal} assesses a model's ability to identify hallucinated content in externally provided captions, rather than directly measuring hallucinations in model-generated outputs, and we acknowledge this inherent gap between discriminative evaluation and generative hallucination. We note that this evaluation paradigm follows from prior work, with established benchmarks~\citep{li2023evaluatingobjecthallucinationlarge, wang2024amberllmfreemultidimensionalbenchmark, videohallucer, vidhallucli2025} similarly adopting discriminative approaches for their capacity for controlled and stable assessment. Crucially, existing hallucination mitigation methods~\citep{huang2024operaalleviatinghallucinationmultimodal, zhou2023analyzing, csr} evaluated on both discriminative benchmarks and open-ended generative tasks consistently show that improvements on the former correlate with reductions in generative hallucinations, supporting discriminative evaluation as a valid proxy for assessing hallucination generation tendencies in VLLMs.

\bibliography{main}
\bibliographystyle{tmlr}

\clearpage
\appendix

\section*{\LARGE Appendix}

\section{Benchmark Construction Details}

\subsection{Dataset Statistics}
\begin{figure}[!ht]
    \centering
    \caption{Distribution of visual instances in \textsc{VidHal} by (Left) public dataset source, categorized by the five temporal aspects, and (Right) temporal aspects and their sub-aspects.}
    \label{fig:source_and_aspect_distribution}
     \begin{subfigure}[t]{0.495\linewidth}
        \includegraphics[width=0.8\linewidth]{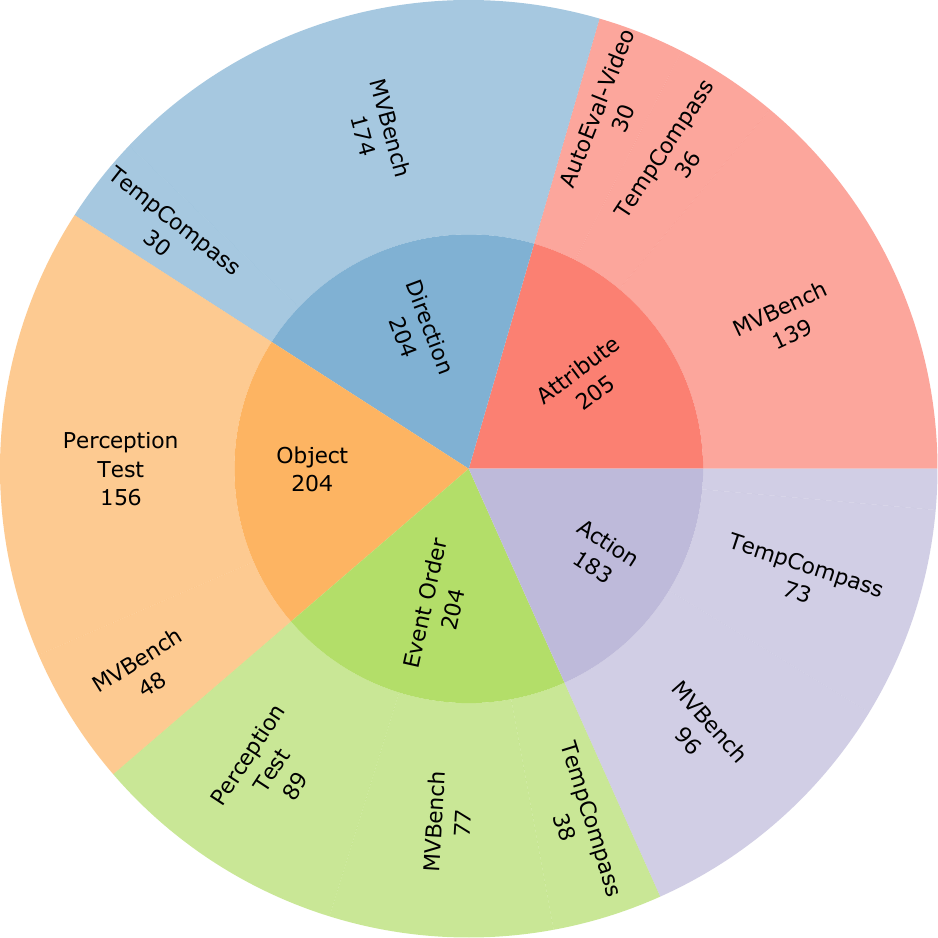}
        \label{fig:data_source_distribution}
     \end{subfigure}
     \hfill
     \begin{subfigure}[t]{0.495\linewidth}
        \centering
        \includegraphics[width=0.8\linewidth]{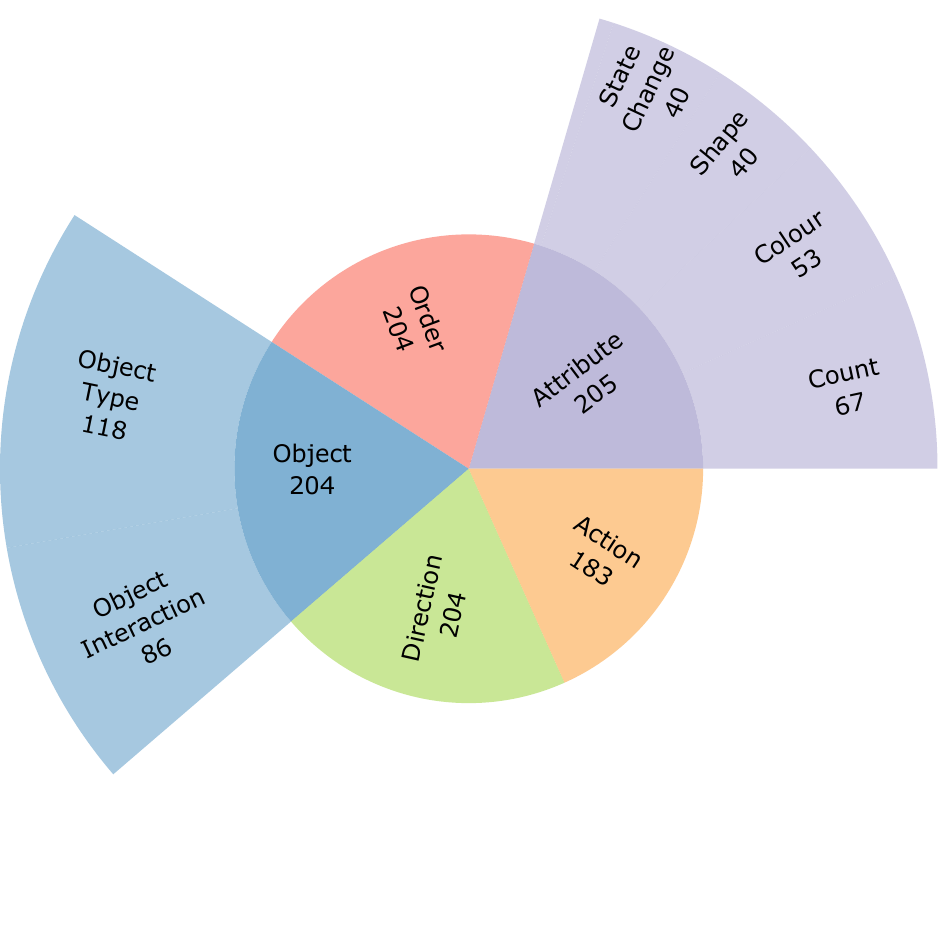}
        \label{fig:sub_aspect_distribution}
     \end{subfigure}
\end{figure}

\begin{figure}
    \centering
    \caption{Distribution of (Left) correct answer options for the MCQA task and (Right) optimal option orders for the caption ordering task.}
    \label{fig:ground_truth_distribution}
    \includegraphics[width=0.6\linewidth]{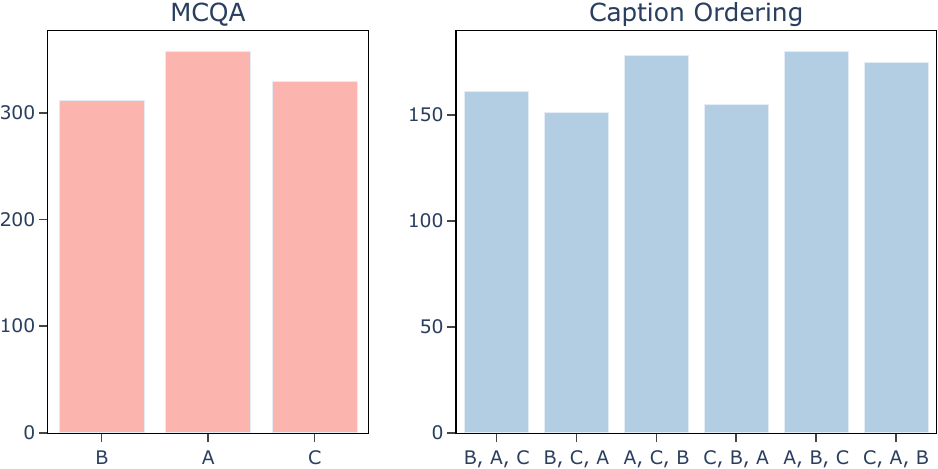}
\end{figure}

\begin{figure}[!tb]
    \centering
    \caption{Distribution of (Left)  caption lengths with an average of 11.2 words, and (Right) duration of videos in \textsc{VidHal} with an average of 15.8s.}
    \label{fig:length_distribution}
     \begin{subfigure}[t]{0.495\linewidth}
         \centering
         \includegraphics[width=\linewidth]{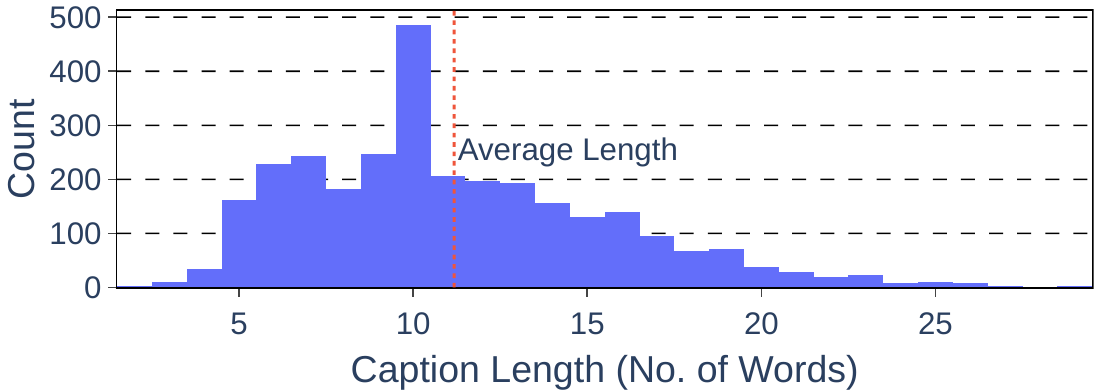}
     \end{subfigure}
     \hfill
     \begin{subfigure}[t]{0.495\linewidth}
         \centering
         \includegraphics[width=\linewidth]{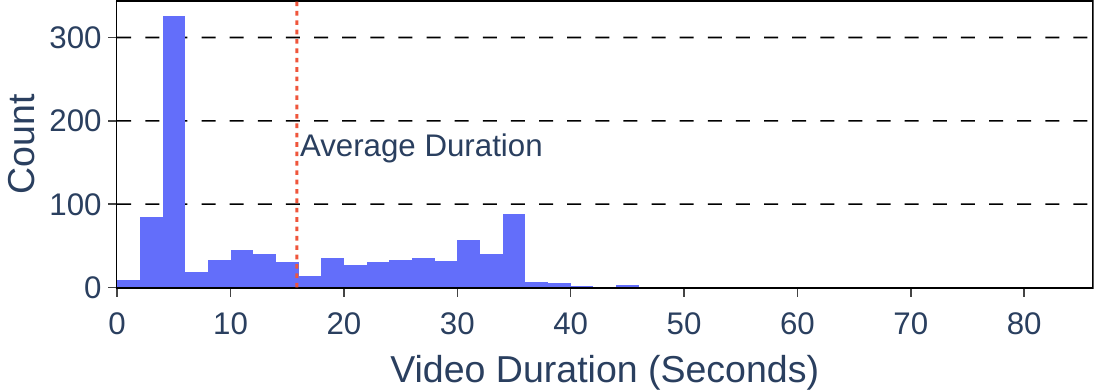}
     \end{subfigure}
\end{figure}

Figure~\ref{fig:source_and_aspect_distribution} presents the distribution of visual instances in \textsc{VidHal} by public dataset sources and temporal aspects. Additionally, Figure~\ref{fig:ground_truth_distribution} further shows the distribution of ground truth answers for the MCQA and caption ordering tasks. One can observe that both temporal aspects and ground truth options are uniformly distributed across our benchmark. The distribution of video caption lengths and video durations is also presented in Figure \ref{fig:length_distribution}.

\subsection{Dataset Development Pipeline}\label{sec:dataset_dev}
\begin{figure*}[!b]
    \centering
    \caption{Specific skills and corresponding questions from the Perception Test dataset chosen for \textsc{VidHal} instance selection, with the matched aspects indicated in brackets.}
    \label{fig:perception_test_categories}
    \includegraphics[width=\linewidth]{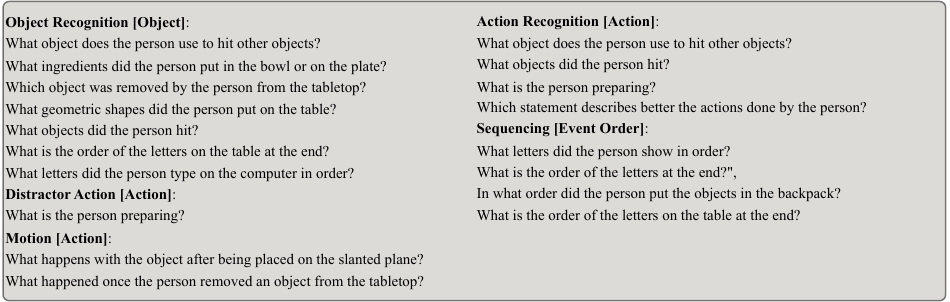}
\end{figure*}

\begin{figure}[!tb]
        \centering
        \caption{Evaluation tasks in MVBench aligned with temporal aspects in \textsc{VidHal}, categorized by aspect.}
        \label{fig:mvbench_categories}
        \includegraphics[width=0.6\linewidth]{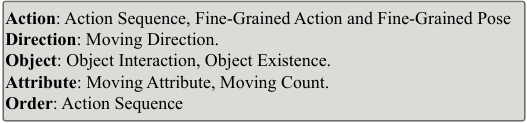}
    \end{figure}

\paragraph{Visual Instance Selection}
To ensure a rich coverage of temporal aspects and visual diversity, we methodically selected video instances from four public datasets: TempCompass~\cite{liu2024tempcompassvideollmsreally}, Perception Test~\cite{pătrăucean2023perceptiontestdiagnosticbenchmark}, MVBench~\cite{li2024mvbenchcomprehensivemultimodalvideo}, and AutoEval Video~\cite{chen2024autoevalvideoautomaticbenchmarkassessing}. Given the unique characteristics of each dataset, we outline the specific guidelines adopted for each dataset below:
\begin{itemize}[leftmargin=10pt]
    \item \textbf{TempCompass} encompasses five temporal aspects: \textit{Action}, \textit{Speed}, \textit{Direction}, \textit{Event Order}, and \textit{Attribute Change}. As most of these aspects align with those chosen to construct \textsc{VidHal}, we retain all video instances except those related to speed. TempCompass includes four evaluation tasks: \textit{MCQA}, \textit{Yes/No QA}, \textit{caption matching}, and \textit{caption generation}. Given the conciseness of captions in the latter two tasks, their information can often be subsumed within the more detailed QA-based annotations. Therefore, we focus exclusively on MCQA and Yes/No QA annotations to create an informative anchor caption.
    \item \textbf{Perception Test} spans various skill and reasoning domains to thoroughly evaluate VLLMs' perception and understanding abilities. Our inspection of these evaluation dimensions reveals alignment between the \textit{semantics}, \textit{physics}, and \textit{memory} skill areas, as well as \textit{descriptive} and \textit{explanatory} reasoning dimensions, with the temporal aspects of action, order, and event order. Accordingly, we limit our video selection in Perception Test to these specific pillars. Additionally, we review the question templates adopted in these areas and select video instances with question-answer pairs that support \textsc{VidHal}'s evaluation objectives. The specific skills and associated questions chosen are detailed in Figure \ref{fig:perception_test_categories}.
    \item \textbf{MVBench} includes twenty video understanding tasks with question-answer pairs designed to challenge the reasoning capabilities of VLLMs. Similar to the Perception Test, we identify the tasks relevant to the temporal aspects in \textsc{VidHal} and focus on collecting videos belonging from these tasks. The specific tasks for each aspect are presented in Figure\ref{fig:mvbench_categories}. We observe that MVBench contains repeated use of certain scenarios across tasks, indicated by similar question templates. To enhance caption diversity and minimize redundancy, we limit the number of examples for each unique scenario. The collected instances cover all five temporal aspects of \textsc{VidHal}.
    \item \textbf{AutoEval-Video} evaluates open-ended response generation in VLLMs through questions with detailed answers across nine skill dimensions. We focus on instances related to the \textit{state transition} area, specifically assessing changes in object and entity attributes. For each instance, we retain the only answers to associated questions as they act as informative, long-form captions for the video.
\end{itemize}

\paragraph{Incorrect Anchor Captions} A minority of videos contain anchor captions misaligned with their content, often due to noisy metadata. Such discrepancies subsequently lead to undesirable hallucinatory captions. To remove such instances, we use BLIP2~\cite{li2023blip2bootstrappinglanguageimagepretraining} to calculate frame-text matching scores across all video frames, selecting the maximum score as the representative video-text alignment score. Examples with incorrect anchor captions typically achieve low alignment scores, which are discarded as noisy instances.\\

\paragraph{LLM-based Caption Generation}
We utilize GPT-4o's~\cite{openai2024gpt4technicalreport} text processing and generation capabilities to generate an anchor caption for each selected video, based on metadata from its original public dataset source. This metadata includes QA-based annotations for TempCompass, Perception Test, and MVBench, along with long-form answers for AutoEval-Video. The anchor caption is subsequently used as input for GPT-4o to generate corresponding hallucinatory captions. 

To ensure the generated hallucinatory captions meet high-quality standards, we employ a detailed prompt adopting the following strategies to guide GPT-4o's output:
\begin{figure}[!b]
    \centering
    \caption{Prompts used for generating the anchor caption from long-form captions.}
    \label{fig:anchor_caption_generation_lc}
    \includegraphics[width=0.6\linewidth]{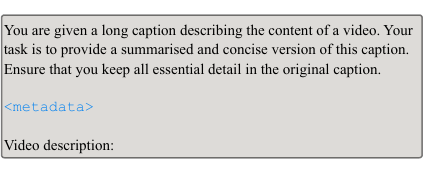}
\end{figure}

\begin{itemize}[leftmargin=10pt]
    \item Aspect-specific definitions which outline the characteristics of each aspect to be varied, prompting GPT-4o to modify anchor captions accordingly.
    \item Caption construction guidelines that define the structure, format, and hallucination levels required for the generated captions.
    \item In-context examples to illustrate the desired form of each hallucinatory caption for each aspect.
\end{itemize}

\begin{figure*}[!tb]
    \centering
    \caption{Prompt for generating aspect-specific hallucinatory captions based on anchor captions and in-context examples.}
    \label{fig:hallucination_caption_generation}
    \includegraphics[width=0.9\linewidth]{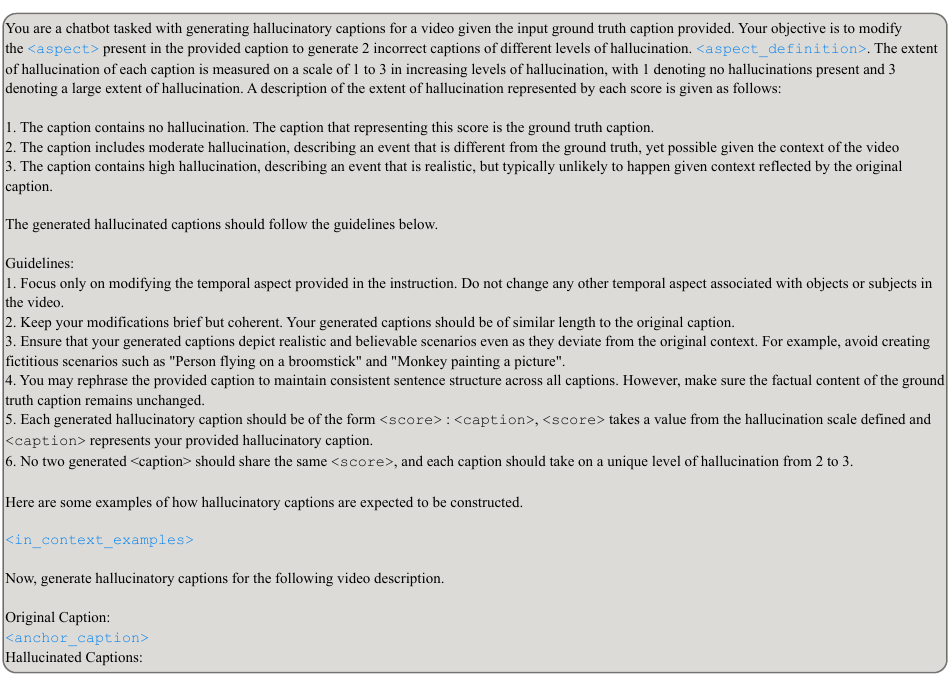}
\end{figure*}

\begin{figure*}[!b]
    \centering
    \caption{Definitions incorporated into the prompt for generating hallucinatory captions for each aspect, with separate definitions provided for each sub-aspect in the \textit{Attribute} aspect.}
    \label{fig:aspect_definition}
    \includegraphics[width=\linewidth]{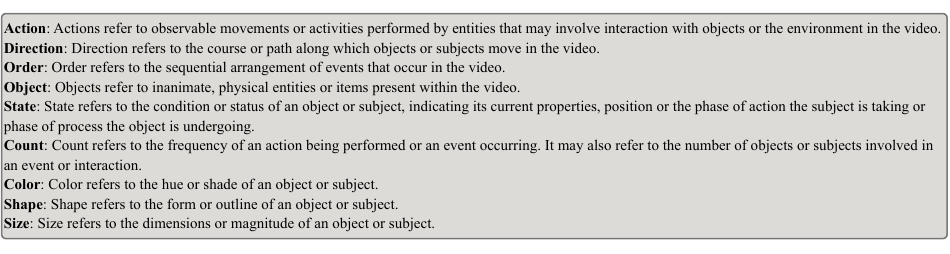}
\end{figure*}

\begin{figure}[!htb]
    \centering
    \caption{(Left) Prompts used for generating the anchor caption, and (Right) in-context examples for the \textit{State} sub-aspect.}
    \label{fig:in_context_size_shape}
     \begin{subfigure}[t]{0.495\linewidth}
         \centering
         \includegraphics[width=0.9\linewidth]{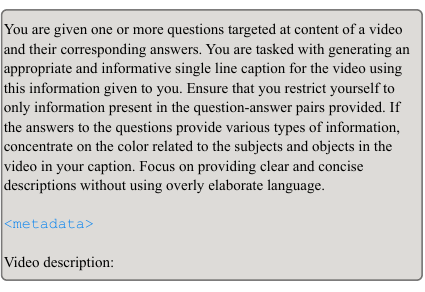}
        \caption{Prompt used for generating the anchor caption from QA-based annotations.}
        \label{fig:anchor_caption_generation_qa}
     \end{subfigure}
     \hfill
     \begin{subfigure}[t]{0.495\linewidth}
         \centering
         \includegraphics[width=0.9\linewidth]{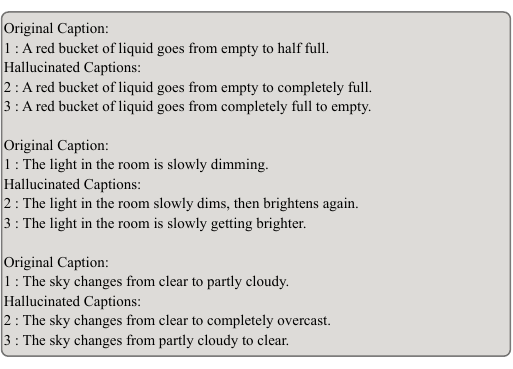}
        \caption{In-context examples for the \textit{State} sub-aspect under the \textit{Attribute} aspect.}
        \label{fig:in_context_state}
     \end{subfigure}
\end{figure}

\begin{figure}[!htb]
    \centering
    \caption{In-context examples for the \textit{Color} (Left) and \textit{Count} (Right) sub-aspects.}
    \label{fig:in_context_color_count}
     \begin{subfigure}[t]{0.495\linewidth}
         \centering
         \includegraphics[width=\linewidth]{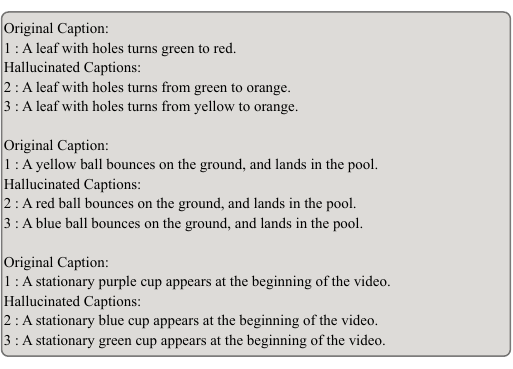}
     \end{subfigure}
     \hfill
     \begin{subfigure}[t]{0.495\linewidth}
         \centering
         \includegraphics[width=\linewidth]{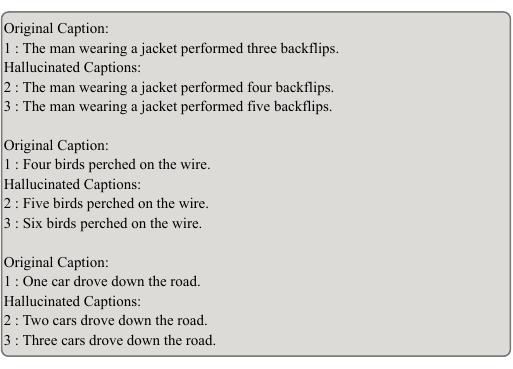}
     \end{subfigure}
\end{figure}

\begin{figure}[!htb]
    \centering
    \caption{In-context examples for the \textit{Object} (Left) and \textit{Event-Order} (Right) aspects.}
    \label{fig:in_context_order_object}
     \begin{subfigure}[t]{0.495\linewidth}
         \centering
         \includegraphics[width=\linewidth]{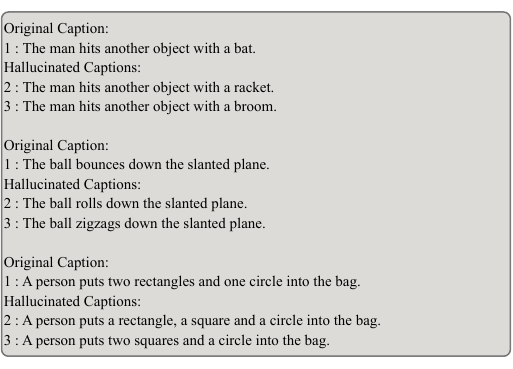}
     \end{subfigure}
     \hfill
     \begin{subfigure}[t]{0.495\linewidth}
         \centering
         \includegraphics[width=\linewidth]{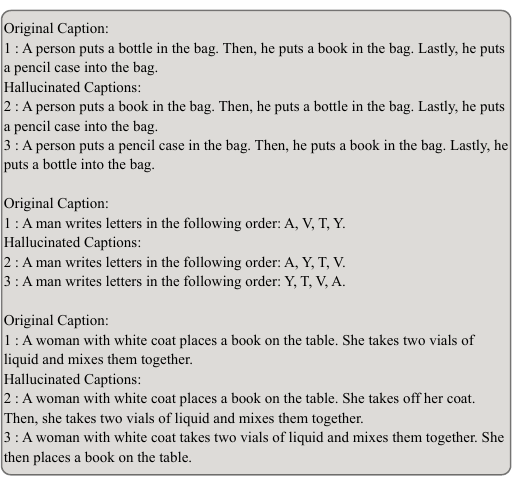}
     \end{subfigure}
     \vspace{-1.4em}
\end{figure}

\begin{figure}[!htb]
    \centering
    \caption{In-context examples for the \textit{Action} (Left) and \textit{Direction} (Right) aspects.}
    \label{fig:in_context_action_direction}
     \begin{subfigure}[t]{0.495\linewidth}
         \centering
         \includegraphics[width=\linewidth]{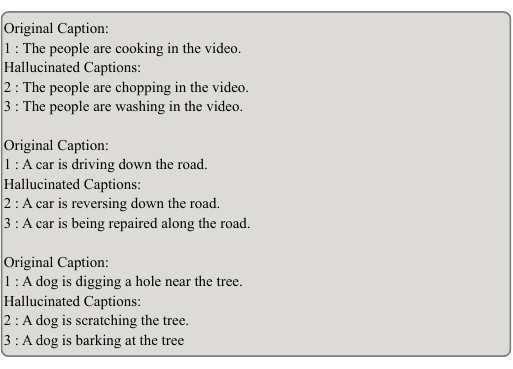}
     \end{subfigure}
     \hfill
     \begin{subfigure}[t]{0.495\linewidth}
         \centering
         \includegraphics[width=\linewidth]{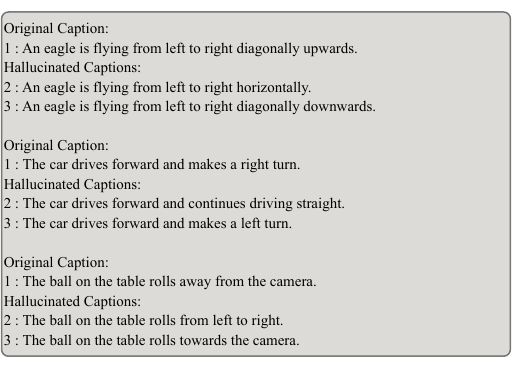}
     \end{subfigure}
     \vspace{-1.5em}
\end{figure}

The prompts for generating anchor and hallucinatory captions are shown in Figures \ref{fig:anchor_caption_generation_lc} to \ref{fig:anchor_caption_generation_qa}, respectively, with definitions for each aspect are provided in Figure \ref{fig:aspect_definition}. Aspect-specific in-context examples are detailed in Figures \ref{fig:in_context_state} to  \ref{fig:in_context_action_direction}. Separate in-context examples are provided for each \textit{Attribute} subaspect of \textit{Shape}, \textit{Size}, \textit{Color}, \textit{Count}, and \textit{State Change} to account for their distinct natures.


\begin{figure}[!htb]
    \centering
    \caption{In-context examples for the \textit{Size} (Left) and \textit{Shape} (Right) sub-aspects.}
    \label{fig:in_context_size_shape}
     \begin{subfigure}[t]{0.495\linewidth}
         \centering
         \includegraphics[width=\linewidth]{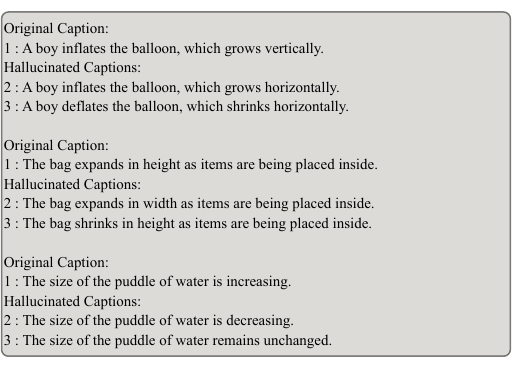}
     \end{subfigure}
     \hfill
     \begin{subfigure}[t]{0.495\linewidth}
         \centering
         \includegraphics[width=\linewidth]{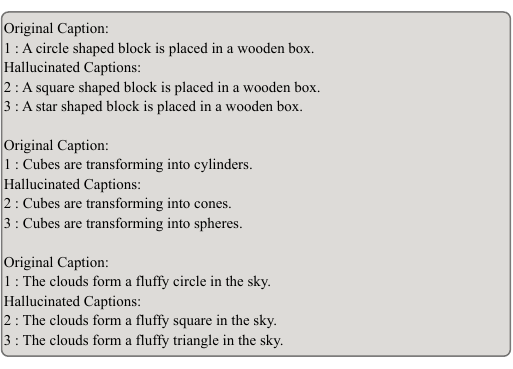}
     \end{subfigure}
\end{figure}

\begin{figure*}[!htb]
    \centering
    \caption{Prompt template for evaluating the quality of generated captions for the GPT-4o, Gemini-1.5 Flash, and LLaMA3 (70B) models.}
    \label{fig:quality_assessment_prompt}
    \includegraphics[width=\linewidth]{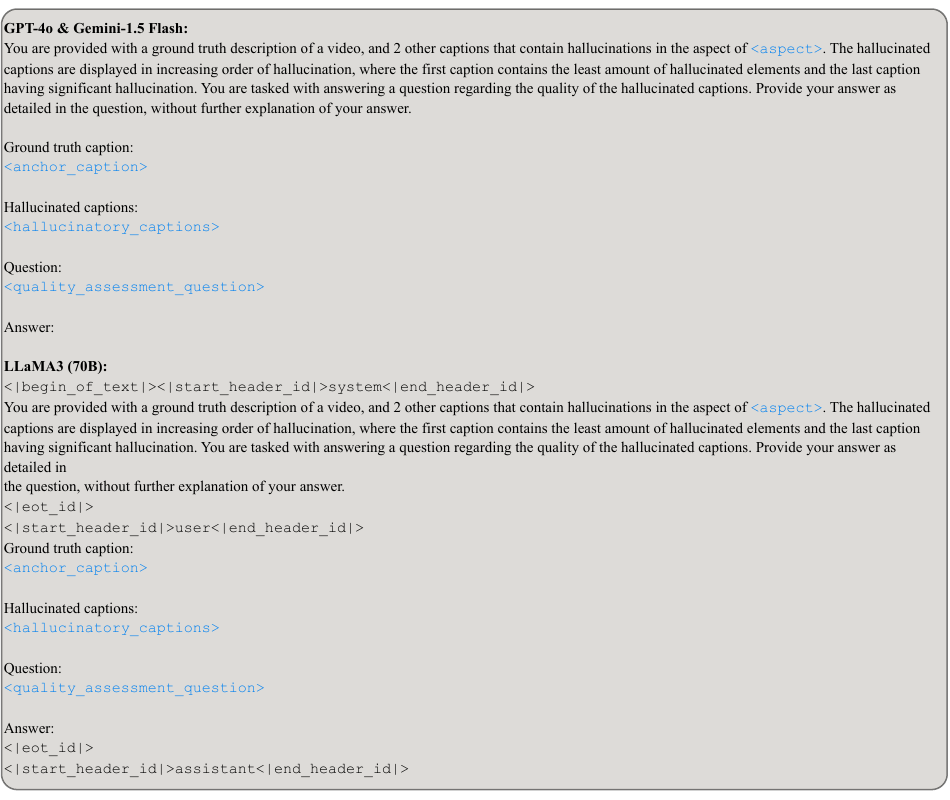}
\end{figure*}
\begin{figure*}[!htb]
    \centering
    \caption{Question prompts for evaluating caption quality based on the three assessment criteria. Prompts with the placeholder \texttt{<option>} are applied individually to the anchor and hallucinatory captions. For question associated with \textit{order quality}, \texttt{<option\_A>} and \texttt{<option\_B>} are replaced with the corresponding hallucinatory caption options shown to the LLMs.}
    \label{fig:quality_assessment_questions}
    \includegraphics[width=\linewidth]{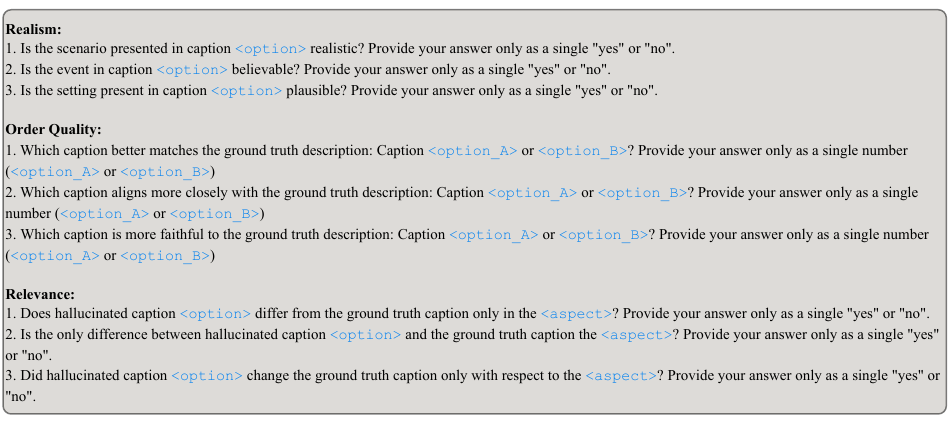}
\end{figure*}

\paragraph{Caption Filtering and Verification} While we adopt carefully designed prompts with detailed severity guidelines and extensive aspect-specific in-context examples to control the generation of hallucinatory captions, GPT-4o may still introduce generative preferences that skew the hallucination ordering or reduce caption quality. To ensure the generated captions meet high quality standards and present reliable hallucination orderings, we adopt a two-stage process of automated filtering followed by manual verification. Specifically, we follow the steps below.

\begin{enumerate}
    \item \textbf{Evaluation Criteria.} For each generated caption set, quality is assessed according to three criteria. \textit{Realism} evaluates whether the hallucinated scenarios described in each caption are plausible and depict realistic events, ensuring that captions do not contain implausible or fictitious content. \textit{Ordering Quality} assesses whether the relative hallucination levels assigned to the captions are appropriate, verifying that captions at higher severity levels do indeed reflect a greater degree of hallucination than those at lower levels. \textit{Relevance} checks that deviations from the anchor caption are confined strictly to the designated temporal aspect, ensuring that no other aspect of the video content is unintentionally altered.

    \item \textbf{Automated Scoring.} To assess each caption set against these criteria, we employ an ensemble of three LLMs: GPT-4o, Gemini-1.5 Flash~\citep{geminiteam2024gemini15unlockingmultimodal}, and LLaMA2 (70B)~\citep{llama3} to enhance robustness against individual model biases. Each LLM independently evaluates the caption set against each criterion using binary questions, assigning a score of 1 for a positive response and 0 otherwise. Details of the prompt templates and criterion-specific questions are provided in Figures~\ref{fig:quality_assessment_prompt} and~\ref{fig:quality_assessment_questions}.

    \item \textbf{Score Aggregation and Filtering.} For each criterion, the binary scores are averaged across the three LLMs in the ensemble. Instances where this average does not meet a majority consensus threshold of $0.66$ (\ie, where fewer than two of the three LLMs assess the criterion as satisfied) are discarded. The per-criterion scores are then summed to produce an overall quality score per instance.
\end{enumerate}
 
\noindent\textbf{Manual Verification.} Following automated filtering, a final manual screening pass is conducted by the authors to verify that the captions and their hallucination orderings are contextually appropriate and consistent with the associated video content. Any residual misalignments, such as captions that are technically plausible but ambiguous in context, are discarded at this stage. From the remaining instances, we retain the top 1,000 examples with the highest overall quality scores, while ensuring a balanced distribution of examples across temporal aspects to construct \textsc{VidHal}.

\FloatBarrier

\subsection{Additional Dataset Examples}
\begin{figure*}[!htb]
    \centering
    \caption{Qualitative examples of video instances and their corresponding generated captions in the \textsc{VidHal} Benchmark, across the five temporal aspects.}
    \label{fig:dataset_examples}
    \includegraphics[width=\linewidth]{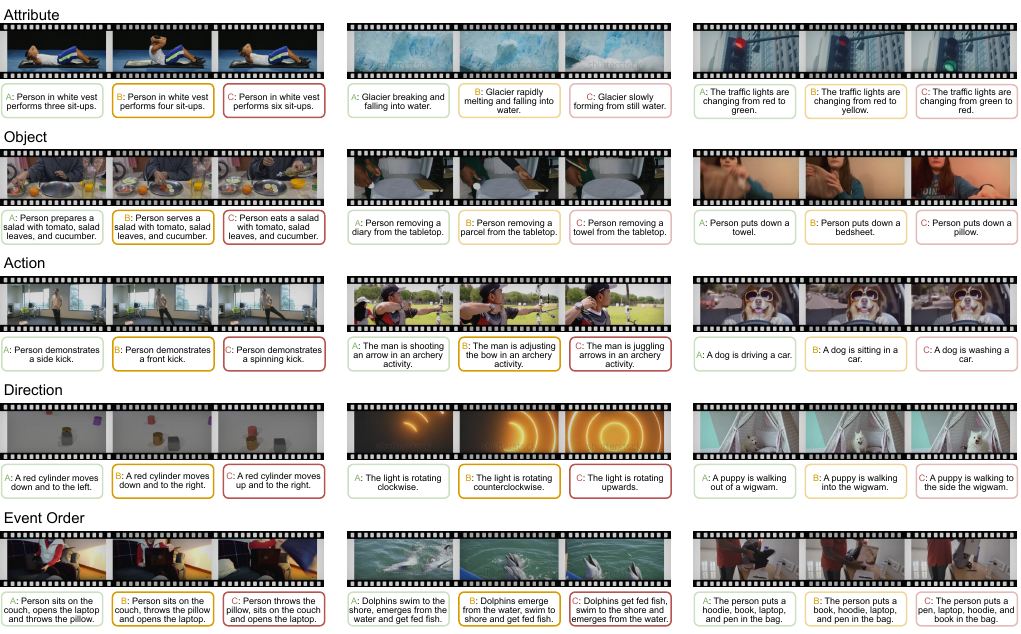}
\end{figure*}

We provide additional qualitative examples of video instances and their corresponding captions in Figure \ref{fig:dataset_examples} for each of the five temporal aspects. 

\subsection{Hallucination Severity Criteria}
Table~\ref{tab:severity_criteria} presents the operationalized criteria for moderate and high hallucination severity across all temporal aspects and sub-aspects in \textsc{VidHal}, along with illustrative caption examples for each level.

\blue{\begin{table*}[!htb]
  \centering
  \caption{Operationalized criteria for moderate and high hallucination severity levels across all temporal aspects and sub-aspects in \textsc{VidHal}, with illustrative examples in ascending order of severity.}
  \resizebox{\textwidth}{!}{%
  \begin{tabular}{p{1.5cm}p{2.25cm}p{5.0cm}p{5.0cm}p{5.0cm}}
    \toprule
    \textbf{Aspect} & \textbf{Sub-aspect} & \textbf{Moderate Severity (MS)} & \textbf{High Severity (HS)} & \textbf{Examples (GT $\rightarrow$ MS $\rightarrow$ HS)} \\
    \midrule
    Action & --- & The depicted action shares similar dynamics or movement patterns with the ground truth and remains plausible within the video context, but does not accurately represent the actual action. & The depicted action shares little to no similarity in dynamics with the ground truth and is inconsistent with the video context. & \textit{A person is kicking a ball.} $\rightarrow$ \textit{A person is throwing a ball.} $\rightarrow$ \textit{A person is juggling a ball.} \\
    \midrule
    Direction & --- & The described trajectory deviates slightly from the ground truth while preserving the coarse axis of movement. & The described trajectory is diametrically opposed to or entirely deviates from the ground truth axis of movement. & \textit{The ball rolls diagonally downward to the left.} $\rightarrow$ \textit{The ball rolls diagonally downward to the right.} $\rightarrow$ \textit{The ball rolls diagonally upward to the right.} \\
    \midrule
    \multirow{2}{*}{Object} & Object Recognition & The identified object shares spatial features (e.g., shape, color, texture) with the ground truth but belongs to a distinct semantic category. & The identified object differs from the ground truth in both spatial features and semantic category. & \textit{A man folds a blanket.} $\rightarrow$ \textit{A man folds a shirt.} $\rightarrow$ \textit{A man folds a paper plane.} \\
    \cmidrule{2-5}
    & Interaction Classification & The described interaction shares similar movements and dynamics with the ground truth but does not accurately reflect the nature of engagement between entities. & The described interaction is categorically distinct from the ground truth and inconsistent with the depicted scenario. & \textit{The person pushes the box across the floor.} $\rightarrow$ \textit{The person kicks the box.} $\rightarrow$ \textit{The person throws the box.} \\
    \midrule
    \multirow{5}{*}{Attribute} & Color & The described color is perceptually adjacent to the ground truth and may be derived from it. & The described color is perceptually distant from the ground truth and bears no resemblance to it. & \textit{Traffic changes from green to red.} $\rightarrow$ \textit{Traffic changes from blue to red.} $\rightarrow$ \textit{Traffic changes from red to blue.} \\
    \cmidrule{2-5}
    & Shape & The described shape is geometrically similar to the ground truth, differing by at most one side. & The described shape is geometrically distinct, differing by at least two sides. & \textit{A hexagonal block is placed in the box.} $\rightarrow$ \textit{A pentagonal block.} $\rightarrow$ \textit{A triangular block.} \\
    \cmidrule{2-5}
    & Size & The described size change deviates from the ground truth in one dimension while remaining accurate in another. & The described size change deviates in both the direction and dimension of change. & \textit{The bag expands in height.} $\rightarrow$ \textit{The bag expands in width.} $\rightarrow$ \textit{The bag shrinks in width.} \\
    \cmidrule{2-5}
    & Count & The described count differs from the ground truth by exactly one. & The described count differs from the ground truth by at least two. & \textit{Three birds on the wire.} $\rightarrow$ \textit{Four birds.} $\rightarrow$ \textit{Six birds.} \\
    \cmidrule{2-5}
    & State Change & The described state transition shares the same starting or ending condition as the ground truth but differs in one dimension. & The described state transition is opposite to or entirely inconsistent with the ground truth. & \textit{The bucket goes from empty to half full.} $\rightarrow$ \textit{Empty to completely full.} $\rightarrow$ \textit{Completely full to empty.} \\
    \midrule
    Event Order & --- & At least one pair of events retains the correct pairwise relative ordering, while at least one other pair has an incorrect ordering. & The ordering of every pair of events is completely reversed relative to the ground truth. & \textit{Picks up book, places in bag, zips bag.} $\rightarrow$ \textit{Places book in bag, picks it up, zips bag.} $\rightarrow$ \textit{Zips bag, places book in it, picks up book.} \\
    \bottomrule
  \end{tabular}}
  \label{tab:severity_criteria}
\end{table*}}

\section{Human Validation Details}
\subsection{Human Validation Process}
\begin{figure*}[!htb]
    \centering
    \caption{Pipeline for validating the quality of generated caption orders in VidHal. For each instance, human annotators are provided with the video and its associated hallucinatory captions. The annotators then select the caption that best aligns with the video content. The selected response is subsequently checked for consistency with the caption with lower hallucination according to our annotation process.}
    \label{fig:dataset_verification}
    \includegraphics[width=\linewidth]{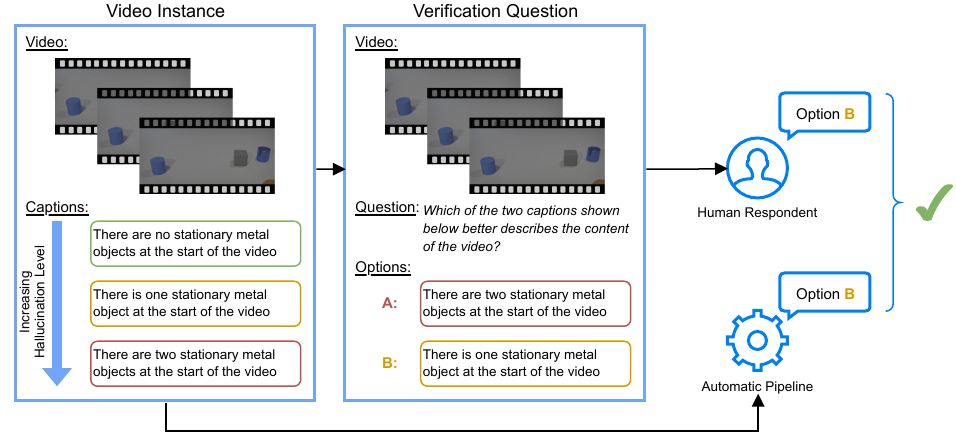}
\end{figure*}

As varying hallucination levels are a distinctive feature of our benchmark, we prioritize validating the robustness of caption ordering produced by our annotation pipeline. Each anchor caption is derived from the original video metadata, making it the most accurate reflection of the video content. Our primary objective is to ensure that the ordering of hallucinatory captions aligns with human judgment. To achieve this, human annotators are shown the video instance along with both hallucinatory captions and are tasked with selecting the caption that better aligns with the video content, as illustrated in Figure \ref{fig:dataset_verification}. Each video instance is reviewed by multiple annotators, with the final human-aligned order determined through a majority vote and compared with our automatically generated order.

\begin{table}[!hb]
    \centering
    \footnotesize
    \caption{Instances where generated caption orders diverge from human preference in quality checks. The agreement score reflects the proportion of respondents who chose our annotated order.}
    \label{tab:misaligned_examples}
    \begin{tabular}{lc}
        \toprule
        Video ID & Agreement Score \\
        \midrule
        \texttt{action\_55} & 0.429 \\
        \texttt{action\_88} & 0 \\
        \texttt{action\_90} & 0.308 \\
        \texttt{action\_118} & 0.200 \\
        \texttt{action\_153} & 0.250 \\
        \texttt{order\_60} & 0.500 \\
        \texttt{order\_109} & 0.154 \\
        \texttt{attribute\_90} & 0.400 \\
        \texttt{attribute\_180} & 0.071 \\
        \texttt{attribute\_192} & 0.188 \\
        \texttt{object\_25} & 0.375 \\
        \texttt{object\_170} & 0 \\
        \texttt{direction\_188} & 0.400 \\
        \bottomrule
    \end{tabular}
\end{table}
\subsection{Analysis of Misaligned Instances}
Table \ref{tab:misaligned_examples} lists video instances that fail to meet the majority agreement threshold established by our annotation process along with their corresponding human agreement scores. To assess the impact of disagreement samples on the \textsc{VidHal} evaluation suite, we evaluated several models on these instances using the full evaluation protocol. Table \ref{tab:noise_analysis} presents the results, demonstrating that model performance on disagreement samples closely aligns with their performance on the complete benchmark, indicating that these instances do not adversely affect the overall evaluation. Upon manual inspection, we found that these disagreement cases predominantly involve visually complex scenarios that are challenging even for some human annotators to verify. Such difficult cases serve as valuable probes for detecting fine-grained hallucinations and distinguishing between state-of-the-art models under perceptually demanding conditions.
\begin{table}[!t]
    \centering
    \caption{Performance of VLLMs on disagreement samples from \textsc{VidHal}.}
    \label{tab:noise_analysis}
    \begin{tabular}{l c c c}
    \toprule
    Model & Accuracy & \multicolumn{2}{c}{NDCG} \\
    \cmidrule(lr){3-4}
     & & Naive & Relative \\
    \midrule
     InternVL2.5 (8B) & 0.769 & 0.491 & 0.816 \\
     Qwen2.5-VL (7B) & 0.923 & 0.814 & 0.777 \\
     Gemini-2.5. (Flash) & 0.846  & 0.931 & 0.834 \\
    \bottomrule
    \end{tabular}
\end{table}

\section{Evaluation Pipeline Details}

\subsection{Model and Inference Hyperparameters}
We provide additional details on the inference and generation settings used across all evaluated models in Table \ref{tab:hyperparameter_config_all}, as well as hyperparameters specific to LlaVA-NeXT-Video models in Table \ref{tab:hyperparameter_config_llava}.
\begin{table*}[!htb]
  \setlength{\tabcolsep}{2pt}
  \centering
  \footnotesize
  \caption{Hyperparameter configuration used in \textsc{VidHal} evaluation across all models.}
  \label{tab:hyperparameter_config_all}
  \normalsize{\begin{tabular}{@{}ll@{}}
    \toprule
    \textbf{Hyperparameter} & \textbf{Value} \\
    \midrule
    \multicolumn{2}{l}{\textit{Data Processing}} \\
    Video Sampling Rate (FPS) & \texttt{30} \\
    \multicolumn{2}{l}{\textit{Generation}} \\
    \texttt{do\_sample} & \texttt{False} \\
    \texttt{temperature} & \texttt{0.0} \\
    \texttt{repetition\_penalty} & \texttt{1.0} \\
    \texttt{max\_new\_tokens} & \texttt{128} \\
    \multicolumn{2}{l}{\textit{Computation}} \\
    Precision & \texttt{FP16} \\
    \bottomrule
  \end{tabular}}
\end{table*}
\begin{table*}[!htb]
  \setlength{\tabcolsep}{2pt}
  \centering
  \footnotesize
  \caption{Model-specific hyperparameters for LLaVA-NeXT-Video models.}
  \label{tab:hyperparameter_config_llava}
  \normalsize{\begin{tabular}{@{}lll@{}}
    \toprule
    \textbf{Hyperparameter} & \textbf{LLaVA-NeXT-Video (7B)} & \textbf{LLaVA-NeXT-Video (32B)} \\
    \midrule
    \texttt{mm\_spatial\_pool\_mode} & \texttt{average} & \texttt{average} \\
    \texttt{mm\_newline\_position} & \texttt{no\_token} & \texttt{grid} \\
    \texttt{mm\_pooling\_position} & \texttt{after} & \texttt{after} \\
    \bottomrule
  \end{tabular}}
\end{table*}

\subsection{Evaluation Task Prompts}

\begin{figure*}[!b]
    \centering
    \caption{Prompt template for the MCQA and relative caption ordering evaluation tasks.}
    \label{fig:evaluation_prompt_mcqa}
    \includegraphics[width=\linewidth]{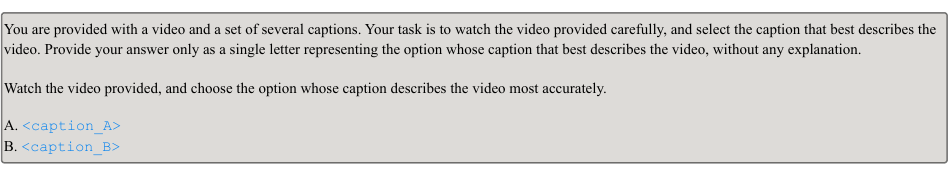}
    \vspace{-1.5em}
\end{figure*}

\begin{figure*}[!tb]
    \centering
    \caption{Prompt template for the naive caption ordering evaluation task.}
    \label{fig:evaluation_prompt_ordering}
    \includegraphics[width=\linewidth]{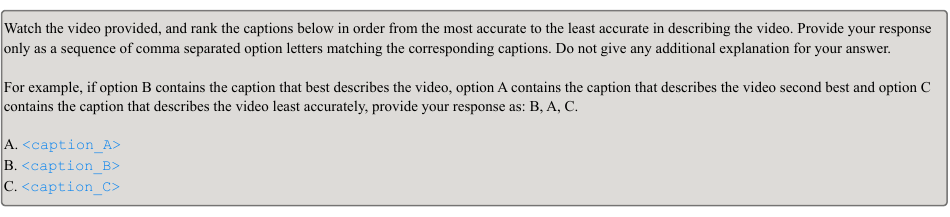}
    \vspace{-1.5em}
\end{figure*}

Figures \ref{fig:evaluation_prompt_mcqa} and \ref{fig:evaluation_prompt_ordering} present the prompts used for the MCQA and naive caption ordering tasks, respectively. The same prompt used for both the MCQA task and the paired questions in the relative caption ordering task. Our manual inspection of these instances reveals that these videos often feature visually complex content, making them challenging even for human annotators.

\subsection{Relative Order Parsing}
\begin{figure*}[!htb]
    \centering
    \caption{Decision tree for determining the final caption order based on VLLM responses to paired questions in the relative caption ordering evaluation task.}
    \label{fig:parsing_logic}
    \includegraphics[width=\linewidth]{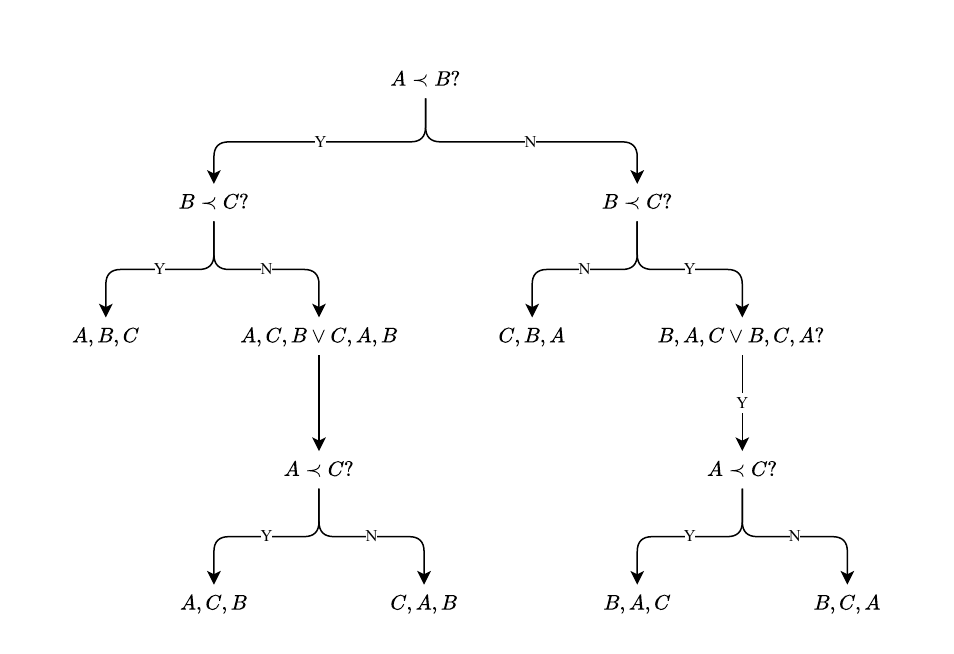}
\end{figure*}

Prompting the VLLM to predict the order of captions based on their hallucinatory level in the relative caption ordering task involves asking a series of paired questions derived from different caption combinations. However, providing the model with all possible pairs at once may result in cyclic and non-transitive orderings. To address this, we present each caption pair to the VLLM in a systematically selected sequence, beginning with two paired questions. The final paired question is presented to the model to resolve inconsistencies if the multiple possible orderings can be derived from the responses to the first two paired questions. The responses across all paired questions presented to the VLLM is then parsed according to the workflow illustrated in Figure \ref{fig:parsing_logic}.

\section{Additional Experiments}

\subsection{Input Order Sensitivity}
\begin{figure}[!htb]
    \centering
    \caption{Distribution of results of VLLMs across varied input caption orders for the three evaluation tasks.}
    \label{fig:input_sensitivity}
    \includegraphics[width=0.75\linewidth]{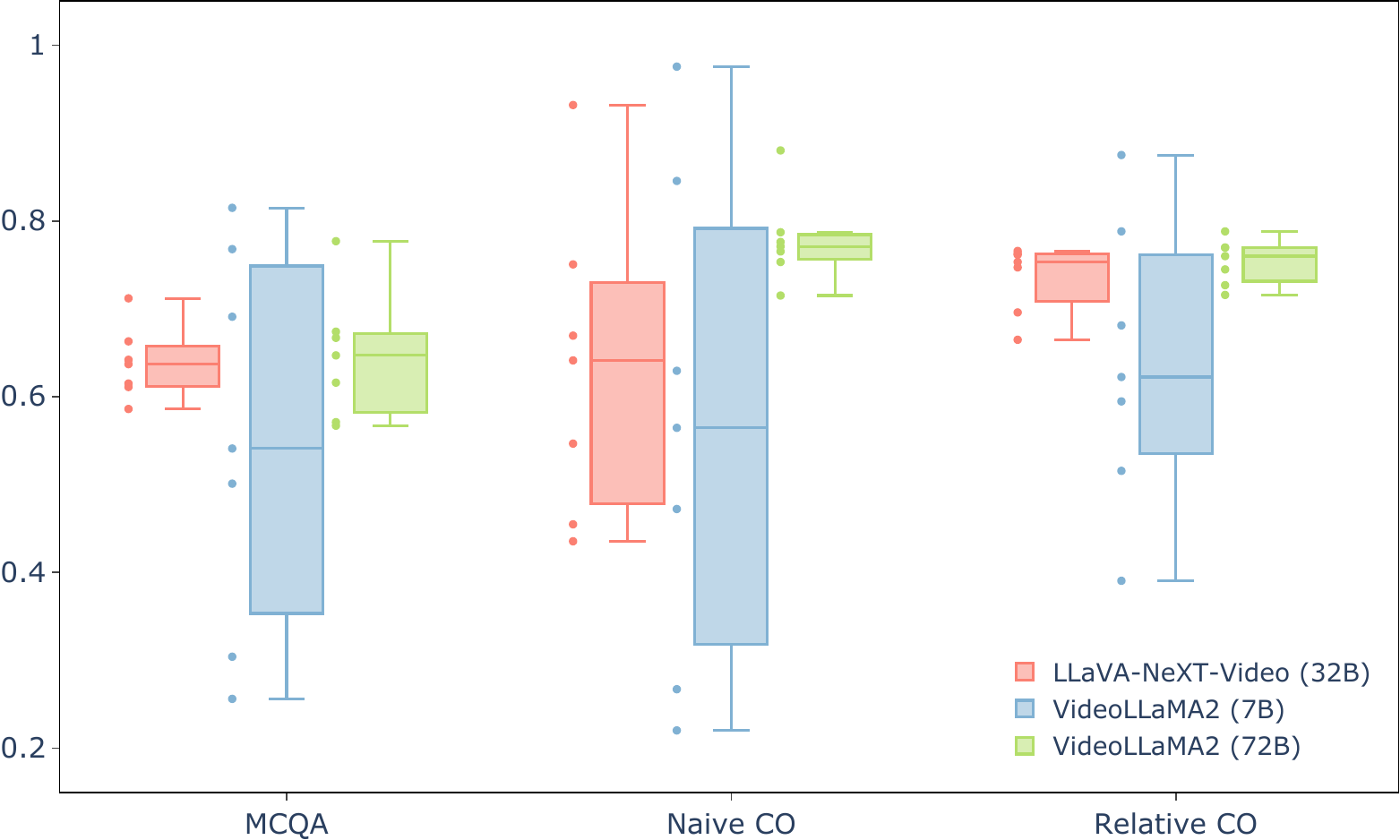}
\end{figure}
To assess the robustness of VLLM responses to the order of displayed captions, we conducted additional experiments by evaluating three VLLMs using a fixed static display order across all instances. We repeated this process across all different permutations of input caption order, presenting the results of these models in Figure \ref{fig:input_sensitivity}. We observe that the performance of these VLLMs is highly sensitive to the order in which captions are displayed, reflected by their varying results across different order permutations. This instability intensifies with smaller model sizes, with VideoLLaMA2 (7B) showing the highest variance in evaluation results and VideoLLaMA2 (72B) the lowest. Our findings suggest that VLLMs may be particularly vulnerable to input caption order, potentially confounding their performance.

\subsection{Naive Caption Ordering Response Quality}
\begin{figure}[!htb]
    \centering
    \caption{(Top) Invalid response rates across all models. VLLMs with no invalid responses are grouped under \textit{Others}. (Bottom) Regurgitation rates of VLLMs on \textsc{VidHal}. \textit{Random} and \textit{Dataset Statistic} indicate the regurgitation rates of the random baseline and ground truth answers, respectively.
    For both metrics, a lower value indicates better model performance.}
    \label{fig:regurgitation_rate}
    \includegraphics[width=0.75\linewidth]{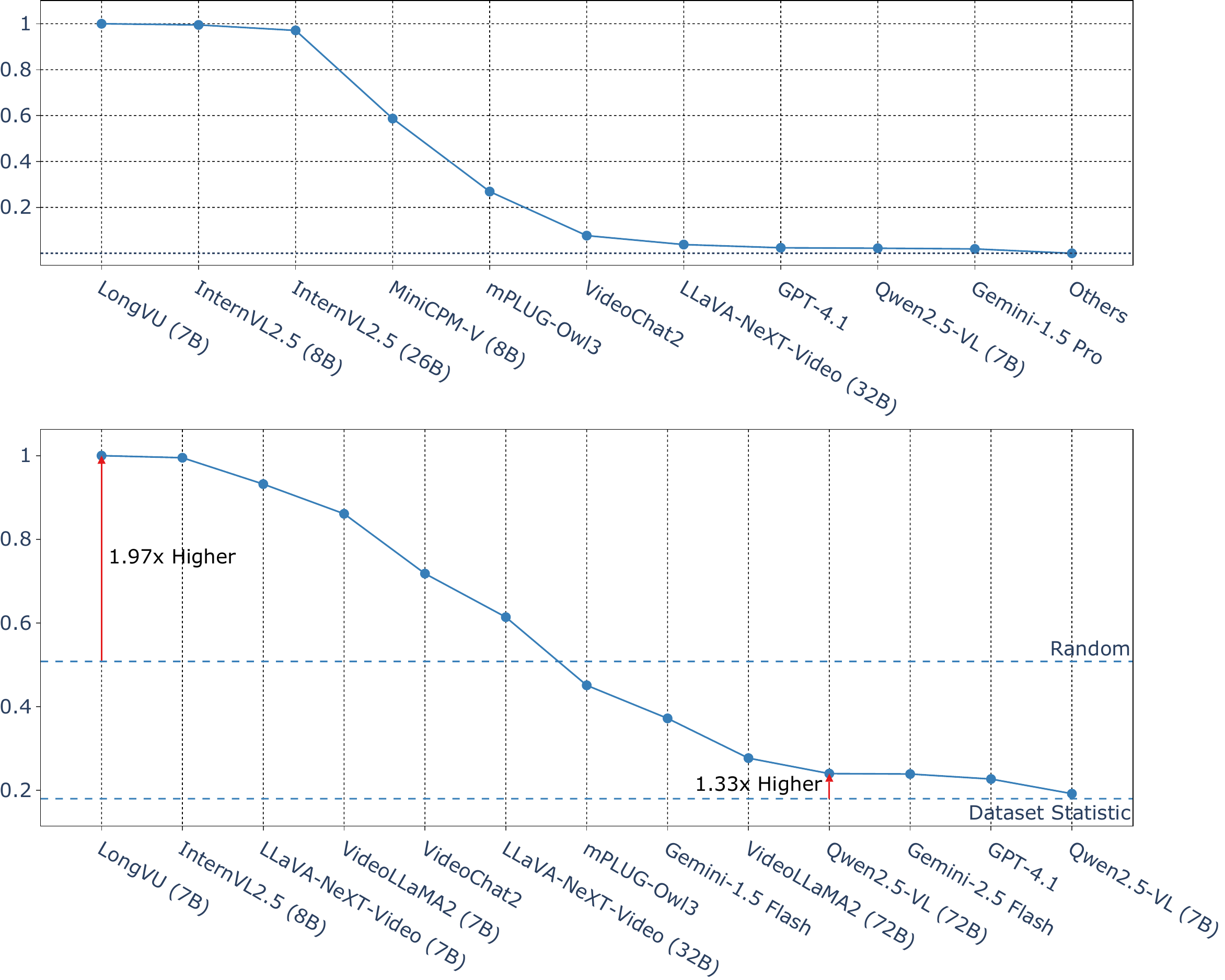}
\end{figure}

To analyze VLLMs' ability to handle naive caption ordering tasks, which possess unique task structures compared to conventional video understanding tasks, we employ two quantitative metrics. Regurgitation Rate (RR) captures the model's propensity to consistently generate identical responses regardless of input, defined as the maximum proportion of instances in \textsc{VidHal} where a specific caption order is predicted across all possible orderings. Invalid Response Rate (IRR) measures the proportion of responses that fail to provide valid caption orders for the naive ordering task. Figure \ref{fig:regurgitation_rate} presents IRR and RR scores for all evaluated models, revealing two key observations. First, many models exhibit high IRR scores, frequently outputting incomplete caption orders (e.g., generating only a single option). Second, despite formulating responses with correct structure, many VLLMs produce identical caption orders regardless of the input video $V^i$, as reflected by high RR scores, a behavior observed even in models performing well on MCQA and relative caption ordering tasks, such as InternVL2.5.

\subsection{Image Prior Reliance - Ablation Study on Video Summarization Algorithm}
We conduct additional single-frame bias experiments using uniform and motion-based sampling strategies with varying clip lengths (1, 2, and 4 frames), with results presented in Tables \ref{tab:image_prior_reliance_uniform} and \ref{tab:image_prior_reliance_motion}. The overlap ratios demonstrate consistency across all three video summarization methods (saliency-based, uniform, and motion-based sampling) for extracting frames $v^i$. In particular, single-frame outputs substantially overlap with full-video inputs regardless of the summarization algorithm employed. These additional results confirm that our single-frame bias study is robust across different frame selection methods, with VLLMs relying on single-frame information for over half of the queries in \textsc{VidHal}.

\begin{table*}[!htb]
  \setlength{\tabcolsep}{2pt}
  \centering
  \footnotesize
  \caption{Overlapping ratios of model predictions under single-frame and full-video inputs for (C)orrect, (I)ncorrect and (O)verall predictions using uniformly sampled frames $v^i$, across multiple frame sampling rates.}
  \label{tab:image_prior_reliance_uniform}
  \normalsize{\begin{tabular}{@{}lcccccccccc@{}}
    \toprule
    & \multicolumn{3}{c}{1 Frame} & \multicolumn{3}{c}{2 Frames} & \multicolumn{3}{c}{4 Frames} \\
    \cmidrule(lr){2-4} \cmidrule(lr){5-7} \cmidrule(lr){8-10}
    \textbf{Model} & \textbf{C} & \textbf{I} & \textbf{O} & \textbf{C} & \textbf{I} & \textbf{O} & \textbf{C} & \textbf{I} & \textbf{O} \\
    \midrule
    VideoLLaMA2 (7B) & 0.674 & 0.708 & 0.700 & 0.781 & 0.798 & 0.794 & 0.846 & 0.829 & 0.833 \\
    LLaVA-NeXT-Video (32B) & 0.680 & 0.570 & 0.620 & 0.735 & 0.649 & 0.688 & 0.831 & 0.706 & 0.763 \\
    \bottomrule
  \end{tabular}}
\end{table*}

\begin{table*}[!htb]
  \setlength{\tabcolsep}{2pt}
  \centering
  \footnotesize
  \caption{Overlapping ratios of model predictions under single-frame and full-video inputs for (C)orrect, (I)ncorrect and (O)verall predictions using motion-based sampled frames $v^i$, across multiple frame sampling rates.}
  \label{tab:image_prior_reliance_motion}
  \normalsize{\begin{tabular}{@{}lcccccccccc@{}}
    \toprule
    & \multicolumn{3}{c}{1 Frame} & \multicolumn{3}{c}{2 Frames} & \multicolumn{3}{c}{4 Frames} \\
    \cmidrule(lr){2-4} \cmidrule(lr){5-7} \cmidrule(lr){8-10}
    \textbf{Model} & \textbf{C} & \textbf{I} & \textbf{O} & \textbf{C} & \textbf{I} & \textbf{O} & \textbf{C} & \textbf{I} & \textbf{O} \\
    \midrule
    VideoLLaMA2 (7B) & 0.521 & 0.495 & 0.515 & 0.558 & 0.507 & 0.519 & 0.670 & 0.653 & 0.657 \\
    LLaVA-NeXT-Video (32B) & 0.634 & 0.550 & 0.558 & 0.658 & 0.546 & 0.597 & 0.675 & 0.563 & 0.614 \\
    \bottomrule
  \end{tabular}}
\end{table*}
\subsection{Hallucination Identification vs. Instruction Following in Naive Caption Ordering}
To disentangle hallucination identification ability from instruction-following proficiency in the naive caption ordering task, we compute NDCG scores restricted to valid responses only and compare them against the overall scores reported in Table~\ref{tab:results}. A response is considered valid if the model produces a complete and parseable caption ordering. Table~\ref{tab:valid_response_ndcg} presents both scores along with the resulting change for all evaluated models. We derive three key observations from this analysis. 

\begin{itemize}
    \item First, models with weaker instruction-following capabilities such as VideoChat2 (Mistral), mPLUG-Owl3, and MiniCPM-V 2.6 experience more substantial score increases when restricted to valid responses, reflecting the significant impact of invalid outputs on their overall NDCG.
    
    \item Second, the InternVL2.5 models exhibit disproportionately large increases attributable to near-complete invalid response rates ($>$95\%), resulting in poor overall NDCG scores. COnsequently, their valid-response scores are computed over very few samples and are therefore subject to high variance, rendering them unreliable reflections of true hallucinatory tendencies. 

    \item Third, and most importantly, relative rankings among top-performing models remain consistent across both evaluation settings. Leading open-source models (Qwen2.5-VL variants, VideoLLaMA2 (72B)) and proprietary models (Gemini-2.5) retain their positions, while models with notable score increases continue to lag behind by a considerable margin.
    
\end{itemize} 

Collectively, these observations affirm the reliability of \textsc{VidHal} and the validity of the experimental insights reported in Section~\ref{tab:results}.

\begin{table}[!ht]
  \centering
  \caption{Naive caption ordering NDCG scores under overall and valid-response-only evaluation settings. Rows marked in bold (InternVL2.5) are flagged due to near-complete invalid response rates ($>$95\%), making their valid-response scores unreliable.}
  \label{tab:valid_response_ndcg}
  \resizebox{0.85\linewidth}{!}{%
  \begin{tabular}{lccc}
    \toprule
    Model & NDCG (Overall) & NDCG (Valid) & Score Change \\
    \midrule
    \multicolumn{4}{@{}l}{\textit{Open-Source Models}} \\
    \midrule
    VideoChat & 0.475 & 0.485 & +0.010 \\
    VideoChat2 & 0.486 & 0.526 & +0.040 \\
    VideoChat2 (Mistral) & 0.503 & 0.630 & +0.127 \\
    VideoChat2 (Phi) & 0.626 & 0.691 & +0.065 \\
    \textbf{InternVL2.5 (26B)} & \textbf{0.475} & \textbf{0.917} & \textbf{+0.442} \\
    \textbf{InternVL2.5 (8B)} & \textbf{0.498} & \textbf{0.726} & \textbf{+0.228} \\
    LLaVA-NeXT-Video (32B) & 0.641 & 0.667 & +0.026 \\
    LLaVA-NeXT-Video (7B) & 0.518 & 0.518 & --- \\
    MiniCPM-V 2.6 & 0.530 & 0.684 & +0.154 \\
    mPLUG-Owl3 (7B) & 0.479 & 0.655 & +0.176 \\
    Qwen2.5-VL (32B) & 0.811 & 0.811 & --- \\
    Qwen2.5-VL (7B) & 0.825 & 0.833 & +0.008 \\
    Qwen2.5-VL (72B) & 0.807 & 0.808 & +0.001 \\
    VideoLLaMA2 (72B) & 0.787 & 0.790 & +0.003 \\
    VideoLLaMA2 (7B) & 0.564 & 0.564 & --- \\
    \midrule
    \multicolumn{4}{@{}l}{\textit{Proprietary Models}} \\
    \midrule
    Gemini-1.5 (Flash) & 0.738 & 0.740 & +0.002 \\
    Gemini-1.5 (Pro) & 0.765 & 0.779 & +0.014 \\
    Gemini-2.5 (Flash) & 0.875 & 0.876 & +0.001 \\
    Gemini-2.5 (Pro) & 0.876 & 0.876 & --- \\
    GPT-4.1 & 0.845 & 0.845 & --- \\
    GPT-4o & 0.840 & 0.840 & --- \\
    \bottomrule
  \end{tabular}}
\end{table}

\subsection{Prompt Sensitivity in Naive Caption Ordering}
To assess whether benchmark conclusions are robust to prompt wording, we evaluated five models across five prompt variations on the naive caption ordering task. We present the results in Table~\ref{tab:prompt_sensitivity}. Overall, stronger models exhibit greater robustness: Qwen2.5-VL (7B) shows the lowest standard deviation ($\sigma = 0.005$), while VideoLLaMA2 (7B) and LLaVA-NeXT-Video (7B) exhibit progressively higher variance consistent with their weaker overall performance. Although InternVL2.5 models also display low variance, this stems from poor instruction-following rather than genuine robustness, as evidenced by their high invalid response rates in Figure~\ref{fig:regurgitation_rate}. With the exception of a minor ranking shift at the lower end of the performance spectrum where LLaVA-NeXT-Video (7B) falls below InternVL2.5 in some prompt variants, all relative rankings remain consistent across prompts, confirming the robustness of our benchmark conclusions to prompt formulation. The full prompt variations used in this experiment are detailed in Table~\ref{tab:prompt_variations}.

\blue{\begin{table}[!ht]
  \centering
  \caption{Prompt sensitivity results on the naive caption ordering task (NDCG) across five prompt variations.}
  \label{tab:prompt_sensitivity}
  \resizebox{\linewidth}{!}{%
  \begin{tabular}{lccccccc}
    \toprule
    Model & Prompt 1 & Prompt 2 & Prompt 3 & Prompt 4 & Prompt 5 & Mean & Std Dev \\
    \midrule
    InternVL2.5 (26B) & 0.498 & 0.546 & 0.557 & 0.474 & 0.574 & 0.530 & 0.037 \\
    InternVL2.5 (8B) & 0.475 & 0.512 & 0.501 & 0.471 & 0.551 & 0.502 & 0.029 \\
    LLaVA-NeXT-Video (7B) & 0.518 & 0.550 & 0.366 & 0.439 & 0.377 & 0.450 & 0.074 \\
    Qwen2.5-VL (7B) & 0.825 & 0.830 & 0.836 & 0.839 & 0.834 & 0.833 & 0.005 \\
    VideoLLaMA2 (7B) & 0.564 & 0.647 & 0.606 & 0.642 & 0.527 & 0.597 & 0.046 \\
    \bottomrule
  \end{tabular}}
\end{table}}

\blue{\begin{table*}[!ht]
  \centering
  \caption{Prompt variations used in the naive caption ordering prompt sensitivity study. Each prompt consists of a system instruction (S), a main task instruction (M), and a formatting hint (H). Prompt 1 indicates the original prompt used in the experiments in our paper.}
  \label{tab:prompt_variations}
  \resizebox{\textwidth}{!}{%
  \begin{tabular}{cp{4.5cm}p{6cm}p{5cm}}
    \toprule
    \textbf{Prompt} & \textbf{System Instruction (S)} & \textbf{Main Instruction (M)} & \textbf{Formatting Hint (H)} \\
    \midrule
    1 & You are provided with a video and a set of several captions. Your task is to order the captions from most to least relevant based on their alignment with the video content. Provide your answer without any further explanation. & Watch the video provided, and rank the captions below from the most accurate to the least accurate in describing the video. Provide your response only as a sequence of comma-separated option letters. Do not include any additional explanation in your answer. & For example, if option B best describes the video, option A describes it second best, and option C describes it least accurately, provide your response as: B, A, C. \\
    \midrule
    2 & You are given a video and a set of candidate captions. Your role is to evaluate how accurately each caption reflects the video content and rank the captions accordingly, from best to worst. Return only your ranking with no accompanying explanation. & Having viewed the video, rank the captions below in descending order of accuracy. Your response should consist solely of a comma-separated sequence of option letters, from the most to the least accurate caption. No further elaboration is required. & For instance, if option A is the most accurate, option C is second, and option B is least accurate, your response should be: A, C, B. \\
    \midrule
    3 & You are a multimodal evaluation system. Your objective is to assess the degree of descriptive alignment between a set of candidate captions and the visual content of a provided video, producing a ranked ordering from highest to lowest alignment. & Upon viewing the video, rank the captions presented below according to their descriptive accuracy relative to the observed content. Your response must be a comma-separated sequence of option identifiers ordered from the most to the least accurate. No supplementary explanation is to be included. & By way of illustration, a response indicating option A as most accurate, option C as second, and option B as least accurate would be expressed as: A, C, B. \\
    \midrule
    4 & You will be presented with a video alongside several descriptive captions. Your objective is to determine the extent to which each caption correctly characterises the video, and to produce a ranking from the most to the least accurate. Provide only the ranking in your response. & After watching the video, arrange the captions below in order of descriptive correctness, beginning with the caption that most precisely reflects the video content and concluding with the least precise. Express your ordering as a comma-separated list of option letters, without any additional commentary. & To illustrate, if option C most precisely reflects the video, followed by option B and then option A, your response should read: C, B, A. \\
    \midrule
    5 & You are provided with a video and multiple candidate captions. Your task is to rank the captions based on their fidelity to the video content, ordering them from the most to the least faithful description. Output only the final ranking. & View the video and order the captions below from the one that most faithfully describes the video to the one that does so least faithfully. Submit your answer as a comma-separated sequence of option letters in ranked order, and include no other content in your response. & As an example, if option B is the most faithful description, option A is second, and option C is least faithful, your answer should be: B, A, C. \\
    \bottomrule
  \end{tabular}}
\end{table*}}

\end{document}